# Frozen Judges, Moving Agents: Version-Dependent LLM-Judge Error and the Limits of Judge-Assisted Agent Evaluation

**Jiapeng Li**

## Abstract

Language-model judges compare agent upgrades with their predecessors, but a fixed judge can make version-dependent mistakes. We analyze 35 public coding-agent submissions (20 prespecified version pairs on 250 SWE-bench Verified issues), two customer-service agents (155 tau-bench tasks), and 1,106 expert-labeled AgentRewardBench trajectories. An upstream outage left three judges for the primary SWE-bench analysis (8,743 aligned cells); the fourth is descriptive. All three coding-agent judges and all four tau-bench judges reject task-conditioned error invariance after multiplicity adjustment. On SWE-bench, 32 of 60 judge-by-pair units have a detectable differential comparison component; eight judge-only intervals declare upgrades that execution-based intervals cannot establish, despite rank correlations of 0.71-0.79. In tau-bench, one judge reverses a nine-point reference-reward gap by penalizing a procedural habit the reward ignores. False acceptance of failed coding patches rises with agent capability conditional on task and execution outcome, while a task-solvability prediction reverses sign across domains. A separately fixed post-submission OpenHands follow-up on the same 250 issues (eight configurations, 1,981 three-judge cells) reproduces the capability/false-acceptance association (mean Spearman +0.944, exact p=0.000099) and decreasing Youden contrast (mean -0.937, p=0.000397); this is observational, not a new-task replication. Transporting old-version calibration raises SWE-bench comparison error from 3.8 to 19.5 points, with 24.6% undefined bootstrap ratios. A paired audit saves only 5% in interval width at 80 labeled tasks. A randomized self-report test is negative (three adjusted p-values=1.0). These results favor paired audits of current outputs over judge-only release decisions or transported calibration; independent human patch review remains pending.

## 1 Introduction

Software teams that build language-model agents increasingly decide whether to ship a new agent version by comparing it with the current one under an automatic evaluator, most often another language model acting as a judge. The practice rests on an assumption that is rarely stated: if the judge is held fixed, a difference in judged success between two versions reflects a difference in their actual success. Monitoring tools reinforce this assumption. They re-score a frozen set of human-labeled outputs to detect changes in the judge (Li, 2026), and calibration methods estimate a judge's sensitivity and specificity on labeled outputs of one system and reuse them for another (Lee et al., 2026).

The assumption fails whenever the judge's error rates depend on which agent produced the trajectory. Measurement-error theory has long distinguished *non-differential* misclassification, which attenuates a comparison toward zero, from *differential* misclassification, which can bias it in either direction (Bross, 1954; Copeland et al., 1977). For language-model evaluation, Dorner, Nastl, and Hardt (2025) showed that model-dependent judge bias can reverse rankings and bounds the labels any debiasing method can save; Fiedler (2026) showed that sharing one calibration set across compared models forces a 1/J amplification of the resulting bias; and several agent benchmarks report judge error that varies across agents (Lù et al., 2025; Xue et al., 2025; Zhuge et al., 2025; Huang, 2026; Bodhwani et al., 2026; Advani, 2026). The phenomenon is therefore not new, and we do not claim to discover it.

What remains unclear is how much it matters for the decision agent developers actually make: comparing a new version with its predecessor, on benchmarks whose outcomes can be verified by execution, when versions are close in quality. Three practical questions follow. How often does a frozen judge reach a different release decision than execution-based evaluation would? Does calibrating the judge on the previous version's labeled outputs repair the comparison? And when labels are scarce, how many does a judge save for a *paired* comparison, as opposed to estimating a single success rate?

We answer these questions with a prospectively planned study on three datasets whose reference labels come from execution or expert annotation. The primary dataset contains the graded patches of 35 publicly released SWE-bench Verified submissions of a single agent scaffold, mini-SWE-agent, which together contain 20 pre-specified pairs of successive public configurations — model upgrades, reasoning-effort and scaffold changes, and variants. Four judge models from three providers were queried on 250 held-out tasks per agent; one provider stopped returning one model's verdicts, so the primary SWE-bench analysis uses the three available judges and reports the fourth on its matched, partial sample. Replications use public τ-bench trajectories with database-state rewards and the 15 released evaluators of AgentRewardBench with expert labels.

After those results were known and the original manuscript had been submitted, we fixed a **separate post-submission validation** using eight publicly graded OpenHands configurations on the same 250 SWE-bench issues, plus a descriptive Agentless configuration. It reuses the three available judges and unchanged patch-only prompt (Sections 4.7 and 5.6). This additional cohort tests whether the capability/error association extends beyond mini-SWE-agent, not whether the original version-pair decision rates or calibration errors generalize.

Our contributions are:

1. **Measurement on execution-grounded version updates.** We estimate class-conditional judge error by agent with task-conditioned tests that separate differential error from differences in prevalence and task mix, and we quantify decision disagreement on pre-specified version pairs.
2. **Why transported calibration fails.** An exact decomposition shows that Rogan–Gladen correction with the previous version's error rates removes the attenuation component of the comparison error but divides the differential component by the old version's Youden index, so it can make close comparisons worse (Proposition 2 gives the condition). We measure how often this happens and compare it with what correction noise alone would produce.
3. **How much a judge can save for a paired comparison.** For a marginally non-differential judge whose errors have conditional mean zero, and two versions with equal outcome and error variances, the ideal label-efficiency gain of judge-assisted inference for a paired difference is the gain for a single success rate scaled by $(1 - r)/(1 - r_\varepsilon)$, where $r$ is the task-level correlation of the two versions' outcomes and $r_\varepsilon$ that of the judge's errors (a general form covers unequal variances). We estimate both correlations and combine the result with a finite-pool sample-size approximation into audit-sizing guidance.
4. **A randomized mechanism test.** Showing the judge the agent's own final message is a common design choice. We randomize it within agent–task cells and estimate its effect on false acceptance, and whether that effect differs across agents. The predicted average effect is not observed.

5. **A reproducible evaluation protocol.** We archive code, prompts, hashes, aggregate results, the preregistration and a deviation log, and describe a paired-audit procedure with asymptotically valid intervals and realistic label budgets. The research repository is private at the time of writing, as the data-availability statement explains.

The post-submission evidence in Section 5.6 is reported in addition to, not as part of, the original prospectively planned three-dataset analysis. Its separate private pre-verdict lock occurred after the original results were known.

## 2 Related work

**LLM judges and their biases.** Judge models show position, verbosity and self-enhancement biases (Zheng et al., 2023; Wang et al., 2024; Panickssery et al., 2024), prefer low-perplexity outputs (Wataoka et al., 2024), and can fail to recognize objectively wrong answers (Tan et al., 2025). Agreement with humans on one distribution does not establish accuracy on another (Dorner et al., 2025, Proposition 2).

**Judging agents.** For web agents, AgentRewardBench (Lù et al., 2025) compares 12 judges with expert labels and reports that judges overestimate and rule-based evaluators underestimate success; Online-Mind2Web reports agent-dependent inflation (Xue et al., 2025). Agent-as-a-Judge documents rank reversals for coding agents (Zhuge et al., 2025); execution-free critics of code changes rank agent workflows imperfectly and lean on trajectory features such as the agent's own reasoning (Yadavally et al., 2025; Jain et al., 2025); and verifiers trained on agent trajectories are used to choose among candidate solutions at inference time (Pan et al., 2025; Jain et al., 2025). On $\tau^2$-bench (Barres et al., 2026), GAUGE finds that judge gates promote the lower-reward agent in close pairs and that recalibration does not transfer (Bodhwani et al., 2026) — the release-decision and transport questions we study on version pairs of coding agents — and Advani (2026) characterizes "false success" in which confident closing messages mislead judges. Huang (2026) asks what leaderboard rank differences estimate on SWE-bench, AgentRewardBench and $\tau^2$-bench: most close orderings are statistically unresolved, and the label source can change which system is selected. Its AgentRewardBench analysis includes a task-conditioned permutation reference and finds that calibrating weak verifiers lowers Brier loss even when the evaluated agent is held out; our AgentRewardBench results (Section 5.1) are a replication with a decomposition of the comparison error and simulated null references added.

**Model-dependent bias and calibration.** Dorner et al. (2025) formalize model-dependent judge bias, prove that rankings can be reversed, and bound the sample-efficiency gains of prediction-powered inference by $1/(1 - \rho^2)$. Lee et al. (2026) derive Rogan–Gladen intervals under an explicit invariance assumption, and Collot et al. (2026) recommend Youden's J for judge selection while noting that their analysis assumes error rates that are stable across the compared models. Fiedler (2026) shows that shared calibration amplifies bias by 1/J and yields confidently wrong signs on MMLU-Pro. ANCHOR (Zhou et al., 2026) transports judge calibration to new models conditional on observed features. Our Propositions 1 and 2 restate these classical and shared-calibration results for version pairs, to connect the transport error to the attenuation component; the empirical question is how large the differential component is for successive public versions of agents.

**Evaluators whose error tracks the evaluated policy.** Optimizing a policy against a fixed reward model eventually lowers the true reward while the proxy keeps rising (Gao et al., 2023), and training with human feedback can raise human evaluators' false-positive rate on the trained model's outputs (Wen et al., 2025). Version-dependent judge error is the evaluation-time counterpart: each release changes the outputs the frozen judge must assess.

**Measurement error.** Differential and non-differential misclassification (Bross, 1954; Copeland et al., 1977; VanderWeele and Hernán, 2012), prevalence correction (Rogan and Gladen, 1978) and the non-transportability of sensitivity and specificity across populations (Ransohoff and Feinstein, 1978; Leeflang et al., 2013) are classical topics that apply directly to judged evaluation.

**Label-efficient evaluation and monitoring.** Prediction-powered inference and its tuned variant use a small labeled sample to correct abundant model predictions (Angelopoulos et al., 2023a; 2023b); related design-based estimators correct surrogate labels in downstream analyses (Egami et al., 2023), and active inference chooses which units to label (Zrnic and Candès, 2024). AutoEval applies these ideas to model evaluation (Boyeau et al., 2025), and prediction-powered ranking builds rank sets (Chatzi et al., 2024). Anchor-based monitoring attributes alarms to the system or the judge (Li, 2026), and prediction-powered risk monitoring provides anytime-valid alarms from sparse true labels (Zhang et al., 2026). We extend the efficiency analysis to paired differences.

**Reference standards for agents.** Test-based resolution in SWE-bench is itself noisy (Yu et al., 2025; Wang et al., 2026; Aleithan et al., 2024), and τ-bench rewards can be gamed by degenerate agents (Zhu et al., 2025). We treat test outcomes and database states as reference standards, not ground truth, and discuss the consequences in Section 7.

## 3 Setting and analytic results

### 3.1 Notation and two notions of non-differential error

For task t and agent version v, let $H_v \in \{0, 1\}$ be the reference outcome (tests pass, database state correct, or expert judgment of success) and $Z_v \in \{0, 1\}$ the judge's verdict. The judge's marginal class-conditional error for version v is described by $TPR_v = P(Z_v = 1 \mid H_v = 1)$ and $FPR_v = P(Z_v = 1 \mid H_v = 0)$, with Youden index $J_v = TPR_v - FPR_v$. Let $p_v = E[H_v]$ be the true success rate and $q_v = E[Z_v] = FPR_v + J_v p_v$ the judged rate. For an ordered pair (old o, new n) evaluated on the same tasks, the true difference is $D_H = p_n - p_o$, the judged difference is $D_J = q_n - q_o$, and the comparison error is $e = D_J - D_H$.

Two notions of non-differential error must be kept apart. The judge is *conditionally non-differential* if, for every task t and label h, $P(Z = 1 \mid H = h, \text{task } t)$ is the same whichever version produced the trajectory: its errors may depend on the task and on whether the trajectory succeeded, but not otherwise on the agent. It is *marginally non-differential* if $TPR_o = TPR_n$ and $FPR_o = FPR_n$. Neither implies the other. If the new version solves tasks whose solutions are harder to recognize, its marginal TPR is lower even under a conditionally non-differential judge; conversely, agent-specific errors can cancel in the marginal rates. Our test of differential error (H1) targets the conditional null. Propositions 1 and 2 are identities in the marginal rates, and Section 4.5 splits the comparison error into a *task-mix* component, which a conditionally non-differential judge can produce, and an *agent-specific* remainder.

### 3.2 Attenuation, reversal and transported calibration

**Proposition 1 (attenuation and reversal).** Write $\delta = \Delta TPR \cdot p_n + \Delta FPR \cdot (1 - p_n)$, with $\Delta TPR = TPR_n - TPR_o$ and $\Delta FPR = FPR_n - FPR_o$. Then

$$e = \delta - (1 - J_o) D_H.$$

If the error is marginally non-differential and $J > 0$, $\delta = 0$ and $D_J = J\, D_H$: the judged difference has the true sign and is shrunk by $J \le 1$. If $J_o > 0$, a sign reversal requires $|\delta| > J_o\, |D_H|$.

*Proof.* $D_J = FPR_n + J_n\, p_n - FPR_o - J_o\, p_o$. Substituting $J_n = J_o + \Delta TPR - \Delta FPR$ and rearranging gives $D_J = \delta + J_o\, D_H$, hence $e = \delta - (1 - J_o)\, D_H$. ■

This is the classical decomposition of misclassification bias (Bross, 1954; Copeland et al., 1977) written for a paired difference; we state it to fix notation. The first term is the *differential* component and the second is attenuation. Close version pairs have small $D_H$, so their comparison error is dominated by $\delta$.

**Proposition 2 (transported calibration).** Suppose $J_o > 0$, the new version's success rate is estimated by the Rogan–Gladen correction with the old version's error rates, $\hat{p}_n = (q_n - FPR_o)/J_o$, and the old version's rate is known from its labels. Then the transported comparison error is

$$e_T = \widehat{p}_n - p_n = \frac{\delta}{J_o}.$$

*Proof.* $q_n - FPR_o = \delta + J_o\, p_n$ by the identity above, so $\hat{p}_n - p_n = \delta / J_o$. ■

Transport removes attenuation but divides the differential component by $J_o < 1$. When $|D_H|$ is small relative to $|\delta|$, $|e_T| > |e|$: calibrating on the previous version makes the comparison worse. This is the version-pair form of the shared-calibration amplification of Fiedler (2026). Two finite-sample facts matter for testing it. First, the estimated error rates carry sampling noise that the correction also divides by $J_o$, so transported estimates are noisier than naive ones even when $\delta = 0$. Second, the naive error of a non-differential judge is the attenuation term, which is large for pairs that differ a lot. The ratio of transported to naive error therefore depends on the mix of pairs and can exceed one without any differential error; we calibrate it against simulated non-differential judges (Section 4.5).

**Proposition 4 (attenuation is set by the tasks that differ).** Suppose the judge is conditionally non-differential with task-specific rates, $E[Z_{v,t} \mid H_{v,t}] = FPR_t + J_t\, H_{v,t}$ for both versions, and let $D_t = H_{n,t} - H_{o,t}$ over a pool of $N$ tasks. Then

$$D_J = \frac{1}{N}\sum_{t=1}^{N} J_t D_t = \frac{N_+ J_+ - N_- J_-}{N} = \bar{J} D_H + \mathrm{Cov}_t(J_t, D_t),$$

where $N_+$ and $N_-$ count the tasks the new version gains ($D_t = 1$) and loses ($D_t = -1$), $J_+$ and $J_-$ are the mean Youden indices on those tasks, $\bar{J}$ is the mean of $J_t$ over all tasks, and $\mathrm{Cov}_t(a, b) = (1/N)\, \Sigma_t\, (a_t - \bar{a})(b_t - \bar{b})$. If $J_+ = J_- = J_\Delta$, then $D_J = J_\Delta\, D_H$.

*Proof.* $E[Z_{n,t} - Z_{o,t}] = J_t\, D_t$, because the false-positive term $FPR_t$ is common to both versions of a task; average over tasks and split the sum by the sign of $D_t$. ■

When $J_+ = J_- = J_\Delta$ with $0 \le J_\Delta \le 1$, a conditionally non-differential judge attenuates the comparison by its discriminability on the *discordant* tasks — those that one version solves and the other does not — rather than by its average discriminability; if judging is harder there, the judge compresses version differences more than its overall Youden index suggests. When $J_+ \ne J_-$, even a conditionally non-differential judge can distort or reverse a difference, because the tasks a version gains and the tasks it loses are judged with different accuracy. Either way the distortion belongs to the task-mix component of Section 4.5, not to agent-specific error.

### 3.3 How much can a judge save for a paired comparison?

Prediction-powered inference with an optimally weighted proxy multiplies the effective labeled sample size by $\tau = 1/(1 - \rho^2)$, where $\rho$ is the correlation between the proxy and the target on a labeled unit (Angelopoulos et al., 2023b; Dorner et al., 2025). For a single version's success rate the proxy is Z and the target H; for a paired comparison the unit is a task, the target is $D = H_n - H_o$ and the proxy is $Z_n - Z_o$.

**Proposition 3 (paired-difference efficiency).** Suppose the judge is marginally non-differential, $Z_v = \text{FPR} + J H_v + \varepsilon_v$, and each verdict depends on the outcomes only through its own version's outcome, $E[\varepsilon_v \mid H_o, H_n] = 0$. Let $\sigma^2_v = \text{Var}(\varepsilon_v)$ and $c = \text{Cov}(\varepsilon_o, \varepsilon_n)$. Then

$$\tau_D - 1 = \frac{J^2 \text{Var}(H_n - H_o)}{\sigma_o^2 + \sigma_n^2 - 2c}, \qquad \tau_{L,v} - 1 = \frac{J^2 \text{Var}(H_v)}{\sigma_v^2}.$$

If the two versions have equal outcome variances and equal noise variances,

$$\tau_D - 1 = (\tau_L - 1)\frac{1 - r}{1 - r_\varepsilon},$$

where $r = \text{corr}(H_o, H_n)$ is the task-level correlation of the two versions' outcomes and $r_\varepsilon = \text{corr}(\varepsilon_o, \varepsilon_n)$ that of the judge's errors.

*Proof.* Because $E[\varepsilon_v \mid H_o, H_n] = 0$, $\text{Cov}(Z_n - Z_o, D) = J\,\text{Var}(D)$ and $\text{Var}(Z_n - Z_o) = J^2\,\text{Var}(D) + \sigma^2_o + \sigma^2_n - 2c$. Hence $\rho_D^2 / (1 - \rho_D^2) = J^2\,\text{Var}(D) / (\sigma^2_o + \sigma^2_n - 2c)$, and similarly for a level. With equal variances, $\text{Var}(D) = 2\,\text{Var}(H)(1 - r)$ and $\sigma^2_o + \sigma^2_n - 2c = 2\sigma^2(1 - r_\varepsilon)$. ■

Two versions of an agent succeed and fail on largely the same tasks, so r is typically well above zero, which shrinks the signal available to the paired proxy. The judge recovers efficiency only insofar as its errors on the two versions' trajectories of a task are themselves correlated — for example, because a task's patches are hard to judge whoever wrote them. Whether a judge helps more or less for a comparison than for a level is thus governed by $(1 - r)/(1 - r_\varepsilon)$, which we estimate. With conditionally uncorrelated errors ($r_\varepsilon = 0$) the paired gain is the level gain times $(1 - r)$. Differential error changes two things: it adds bias ($\delta$ above), which a labeled audit removes, and it changes the proxy's correlation with the target, which can rise or fall; Proposition 3 is the non-differential benchmark against which we compare observed efficiencies. It also concerns a single linear proxy; richer proxies, such as the judge's probabilities or several judges' verdicts, can do better (Section 5.2).

**Corollary (judge-only tests).** If $E[Z_n - Z_o \mid H_o, H_n] = \beta (H_n - H_o)$ for some $\beta > 0$, the judge-only paired z-test on N tasks has noncentrality

$$\lambda_J = \frac{\rho_D \sqrt{N}\, D_H}{\text{SD}(D)}, \qquad N_{\text{equivalent}} = N\rho_D^2,$$

that is, the power of a test on $N \rho_D^2$ labeled task pairs. *Proof.* $\text{Cov}(Z_n - Z_o, D) = \beta\,\text{Var}(D)$, so $\rho_D = \beta\,\text{SD}(D)/\text{SD}(Z_n - Z_o)$, and the judge-only noncentrality is $\sqrt{N}\,\beta\, D_H / \text{SD}(Z_n - Z_o)$. ■ The single-rate version is the classical loss of power from misclassification. With differential error, $E[Z_n - Z_o \mid H_o, H_n]$ need not be proportional to $H_n - H_o$, so the judge-only test can be biased — it can reject when $D_H = 0$ — as well as weaker or stronger; we therefore report $N \rho_D^2$ only as a description.

### 3.4 What fixed anchors certify

Anchor-based monitors re-score a frozen set of labeled outputs and alarm when the judge's agreement with the labels changes (Li, 2026). Because the anchor outputs are fixed, the anchor statistic is a function of the judge and the anchors only; by construction it cannot respond to a change in the judge's error that is caused by the *new* version's outputs (Li, 2026, Proposition 2). Anchors certify the stability of the judge, not its validity on the outputs being compared. Section 5 asks how large the resulting blind spot is for successive versions of real agents.

## 4 Study design

The study plan was frozen and committed to a **private** research repository before any confirmatory judge request was sent (`preregistration/01-confirmatory-plan.md`, frozen 2026-09-23 with file hashes), after a discovery pilot on the AgentRewardBench development split (`00-pilot-protocol.md`). Here "preregistered" denotes prospective specification within that repository, **not** an independently time-stamped or publicly registered record in the sense described by Nosek et al. (2018); commit history alone cannot provide that stronger assurance. All deviations are logged in `preregistration/DEVIATIONS.md` and summarized in Section 4.6.

### 4.1 Datasets

**SWE-bench Verified (primary).** SWE-bench asks an agent to resolve a real GitHub issue, and a submission is graded by running the repository's tests (Jimenez et al., 2024); the Verified subset contains 500 human-screened instances (Chowdhury et al., 2024). mini-SWE-agent is a minimal bash-only scaffold from the SWE-agent project (Yang et al., 2024) that has been run with many models. We collected all 48 mini-SWE-agent submissions on the SWE-bench Verified leaderboard and applied a fixed inclusion rule: outcomes recorded for all 500 instances (41 entries record per-instance outcomes; one of them covers only 441), a recorded resolve rate within one point of the reported rate (two entries fail), and public artifacts — the graded patch from the evaluation logs or, for entries without logs, the submission recorded in the trajectory — for at least 95% of sampled tasks (three entries fail). Thirty-five agents qualified. On a random sample of 242 agent-task cells with both sources, the evaluation-log patch and the trajectory submission were identical up to surrounding whitespace, after we corrected a character-decoding error in our own download code. Tasks: 250 confirmation instances drawn once with a fixed seed and stratified by repository, disjoint from 40 development instances used only to test prompts and parsing. The reference label is the recorded test-based resolution.

**τ-bench.** Public historical trajectories of GPT-4o and Claude 3.5 Sonnet (new) on the airline and retail domains (Yao et al., 2025), trials 0–3, with the environment reward as reference label; 155 confirmation tasks after reserving 10 for development.

**AgentRewardBench.** The official test split (300 tasks, four agents, 1,106 trajectories with a primary expert annotation of success or failure) and the 15 released evaluators, including the benchmarks' rule-based evaluators (Lù et al., 2025). No new model calls are made. Unsure annotations are excluded.

### 4.2 Judges and prompts

For SWE-bench and τ-bench, four judge models from three providers — GPT-5 mini and GPT-6 Sol (OpenAI), Gemini 3.5 Flash (Google) and Claude Opus 5.5 (Anthropic) — were queried through an OpenAI-compatible gateway with provider-default decoding, one sample per request. The SWE-bench judge sees the repository name, the issue and the graded patch and returns a verdict (RESOLVED or NOT_RESOLVED) with a probability; the τ-bench judge sees the domain policy and the rendered conversation, including tool calls and results, and returns SUCCESS or FAILURE. Prompts are in Appendix A. A verdict that cannot be parsed is re-sampled once; if still invalid, the primary analysis codes it as negative and a sensitivity analysis excludes it. Requests were issued in a fixed random order over tasks and agents, with the four judges of a trajectory adjacent, so time-limited shortfalls would be dispersed across agents and tasks. This does **not** guarantee random missingness when upstream errors depend on the prompt: the first 15 Gemini failures clustered on nine issues, and seven remained on three issues after a ten-attempt targeted retry pass. Claude Opus 5.5 stopped returning verdicts after 7,653 of 8,750 SWE-bench cells (87.5%); GPT-5 mini, Gemini 3.5 Flash and GPT-6 Sol produced error-free responses on 8,750, 8,743 and 8,750 cells, respectively; one GPT-5-mini response was unparseable and was coded negative under the frozen rule. The three-judge primary sample was frozen at 8,743 aligned agent-task cells on 2026-09-25. The Claude outage happened before SWE-bench verdicts were read. Rather than mislabel a replacement judge as Claude or discard the other judges' valid cells, we use the three available judges on the same complete (agent, task) cells as the modified SWE-bench primary analysis. Claude is retained only in a separately labeled four-judge matched-subset sensitivity. For the randomized mechanism experiment each judge is analyzed on its own complete three-arm (agent, task) cells. The original four-judge completeness and global-Holm rules are thus not literally fulfilled (Section 4.6).

### 4.3 Version pairs

Before any confirmatory judgment existed, we fixed 20 ordered pairs of adjacent releases within a product line: 12 model upgrades (for example GPT-5 → GPT-5.1 → GPT-5.2, Claude Opus 4 → 4.5 → 4.6, GLM-4.5 → 4.6 → 5), one reasoning-effort change, two scaffold upgrades with the model fixed, three combined scaffold-and-configuration changes, and two size or specialization variants (Appendix B). All 595 agent pairs are analyzed secondarily.

### 4.4 Hypotheses

The preregistered hypotheses and decision rules were:

- **H1 (differential error).** Judge verdicts depend on the agent after conditioning on task and reference label. Test: permutation of agent labels within (task, label) strata, with the between-agent variance of acceptance rates as statistic; supported for a dataset if at least half of its LLM judges reject after Holm adjustment.
- **H2 (biased version comparisons).** The differential component of the comparison error, $e_d = D_J - J_{pool} D_H$, differs from zero more often than chance, and judge-only release decisions (better, worse, inconclusive by 95% paired intervals) disagree with reference-label decisions.
- **H3 (transport fails).** Rogan–Gladen transport from the old version leaves more than half of the naive comparison error (lower 95% bound of the error ratio above 0.5).
- **H4 (audits).** Paired audits with prediction-powered intervals have valid coverage where judge-only intervals do not, and the ideal efficiency gain for paired differences is smaller than for levels.
- **H5 (self-report mechanism).** Adding the agent's final message to the judge's input raises false acceptance, unevenly across agents (SWE-bench, 21 agents whose trajectories expose a final message, 100 tasks, all four judges).

After the AgentRewardBench analysis, and before any SWE-bench or τ-bench verdict was read, we registered two further directional hypotheses suggested by that analysis (Section 5.1):

- **H6 (capability gradient).** Across SWE-bench agents, the within-task FPR contrast rises and the within-task Youden contrast falls with the agent's reference resolve rate. Test: mean over judges of the per-judge Spearman correlations, with a permutation test that permutes resolve rates across agents; supported if both correlations have the predicted sign with $p < 0.05$.
- **H7 (registered as the "near-miss gradient"; here the solvability gradient).** On SWE-bench, among failed patches, the judge's acceptance rate increases with the share of the other 34 agents whose patch for the same task was resolved. Test: per judge, the slope of a linear probability model with a 95% task-cluster bootstrap interval; supported if the slope is positive with an interval excluding zero for at least three of the four judges. Because the share of other agents that solved a task measures the task, not how close the focal patch came to a solution, we call the association the solvability gradient and treat "near miss" as one possible interpretation.

An internal, model-assisted methodological review of the frozen plan, completed while the confirmatory SWE-bench and τ-bench requests were running and before any of their verdicts had been read, found that several of these rules are weakly informative. The preregistered H1 statistic pools both labels within an agent, so it has power mainly when agents that succeed more often are also accepted more often when they fail. The H3 criterion can be met by a judge with no differential error, because transport amplifies sampling noise and the naive error of a non-differential judge is pure attenuation. And half of the preregistered version pairs have reference differences too small for a binary "significant or not" count to have power. We therefore report, beside every preregistered result, the following analyses, which were logged and committed in the private repository before the remaining data were read: a label-specific H1 statistic and per-pair McNemar tests; the task-mix and agent-specific decomposition of comparison error; calibration of the H2 and H3 summaries against simulated non-differential judges; continuous comparison-error estimates with pooled means and minimum detectable effects for the version pairs; audit coverage by audit size with Student-t intervals and an attenuation-corrected comparator; and a placebo-controlled mechanism experiment.

### 4.5 Statistical analysis

*Resampling.* Tasks are the resampling unit for paired-difference intervals and audits. Intervals use a multinomial task bootstrap (2,000 draws; Efron and Tibshirani, 1993) shared by all agents, pairs and judges, so that pooled statistics respect shared tasks; permutation tests use 9,999 permutations. H1 is adjusted with Holm's (1979) method within each dataset (for the dataset-level rule) and across all judge-by-dataset tests; H2 uses the Benjamini–Hochberg (1995) procedure within judge. The added H6 association instead permutes capability labels across agents, with the same permutation for each judge.

*Differential error.* Besides the preregistered statistic, the label-specific statistic sums, over agent-by-label cells, the squared difference between a cell's acceptances and those expected from its (task, label) strata, divided by the cell size; it is sensitive to departures in either direction and uses the same within-stratum permutation. For τ-bench, where each agent attempted each task four times, an agent's trials on a task with the same label are permuted together. Because trials of one agent on one task can be correlated in ways a permutation does not respect, we also refer the label-specific statistic to a task-level wild bootstrap: each task's vector of observed minus expected acceptances is multiplied by a random sign, which is valid for any within-task correlation as the number of tasks grows. This cluster-robust p-value is the primary H1 evidence for τ-bench, where the preregistered trial-level permutation is reported as a naive sensitivity analysis, and per-agent τ-bench error rates carry task-bootstrap intervals. Per-agent within-task contrasts compare each agent's acceptance rate with the other agents' rate on the same task and label, averaged over tasks with equal weights; they estimate task-conditioned differences in FPR (label 0) and TPR (label 1), free of differences in which tasks the agents solve.

*Comparison error and its components.* For each judge and ordered pair, the comparison error $e = D_J - D_H$ is estimated on the tasks both versions attempted. The task-mix component predicts each version's judged rate from its own outcomes and the judge's acceptance rates for *other* agents' trajectories on the same task and label (pooled rates where no other agent shares the label); its difference from $D_H$ is the error a conditionally non-differential judge would make — attenuation plus the effect of the two versions solving different tasks. The agent-specific component is the remainder. Where no other agent shares a task and label, the pooled rate for that label is used, so the split is a leave-pair-out prediction rather than an identified causal decomposition; we report for each pair the share of trajectories whose task and label are supported by other agents, and the split is undefined for τ-bench, which has only two agents per domain. Appendix C checks these diagnostics by simulation. Where every task has one trajectory per agent, exact McNemar tests compare the two versions' acceptance on tasks both versions fail (a paired FPR test) and on tasks both solve (a paired TPR test). The differential component $e_d$ uses the pair's pooled Youden index, re-estimated in every bootstrap replicate. Release decisions use 95% percentile intervals of the paired differences. The preregistered binomial test of the share of significant differential components treats comparisons as independent although they share agents, tasks and judges; we report it as planned but base pooled inference on the null references below. Continuous $e$ and $e_d$ are reported with intervals for every version pair, and their signed means across judge-pair units have joint task-bootstrap intervals. A per-unit minimum detectable magnitude for $e_d$ at two-sided 5% significance and 80% power is approximated by $(z_{0.975} + z_{0.8})$ times its task-bootstrap standard deviation; it is a planning quantity, not a post-hoc test.

*Null calibration.* Decision disagreement, the fraction of significant differential components, point-sign reversals, the mean agent-specific error and the H3 ratio are compared with their distributions under two simulated non-differential judges (200 simulations each, 500 bootstrap draws per simulation): a *conditional* null that permutes a judge's verdicts among agents within (task, label) strata, preserving task-level judging difficulty, and a *marginal* null that draws each verdict from the judge's overall TPR or FPR independently. Both assume one trajectory per agent and task, so they are not computed for τ-bench.

*Transport.* The old version's TPR and FPR are estimated on all of its tasks and applied to the new version's judged rate; corrected rates outside [0, 1] are clipped, and we report how often. We report the preregistered percentile interval of the transport-to-naive ratio and a basic bootstrap interval. A bootstrap draw in which the old-version Youden index becomes nonpositive for any point-defined pair is omitted *in full*: both errors are averaged over the same fixed set of judge-pair units in every retained draw. We report the fraction of undefined draws, so the interval is explicitly conditional on the estimator remaining defined. The non-differential null simulations use the same positive-Youden point-unit rule and report any undefined simulations.

*Audits.* Audit simulations draw task samples of size 20–160 without replacement from the common tasks of a pair, label both versions, and compare classical, prediction-powered (PPI) and power-tuned (PPI++) intervals for the pool's reference difference, with finite-population corrections and normal or Student-t quantiles, against the judge-only interval a practitioner would compute and an attenuation-corrected judge estimate (judged difference divided by the Youden index estimated on the audited trajectories). The H4(b) Wilcoxon test is descriptive because pairs share agents and tasks.

*Mechanism.* For each judge, arm contrasts are computed within (agent, task) cells on failed patches (FPR) and resolved patches (TPR), tested by sign-flipping all of a task's contrasts together, with agent heterogeneity tested by permuting agent labels within task. The primary test is one-sided on FPR for report versus concurrent control, Holm-adjusted across the three available judges after the outage; any partial Claude result is descriptive and does not enter this family. FPR and TPR arm differences receive 95% task-bootstrap intervals. Changes in Youden's J use a joint bootstrap of both labels' task counts; draws with no observations in a required label are omitted and counted.

*Reference-error sensitivity.* The execution-based resolution label is an imperfect reference. For each of the 35 SWE-bench agents, seed 20260924 samples up to five failed patches accepted by a majority of the three available judges and up to five resolved patches rejected by a majority, plus up to two unanimous judge–test agreements of each label. A separate GPT-5.5 adjudicator sees the issue, the maintainers' reference patch and test changes, the tests that must pass, and the candidate patch, but not the test outcome or any judge verdict (Appendix D). Disagreement with the reference is reported by stratum with Wilson intervals and by agent, with a permutation test for the capability gradient. A *descriptive* error-rate sensitivity assigns each cell a probability that its reference label is wrong from the sampled agent-by-stratum adjudications, shrunk toward its pooled stratum rate with weight five. Both originally failed and originally resolved labels are eligible for correction; cells with unobserved intermediate vote splits interpolate between the two adjacent stratum rates. The resulting expected FPR and FNR use each cell's estimated probability of actual failure or success as a weight. This is not a new gold standard or a design-unbiased correction: the adjudicator can err, and the interpolation and shrinkage are assumptions.

### 4.6 Deviations

All changes to the frozen plan are logged with dates in `preregistration/DEVIATIONS.md` and recorded by private-repository commits; the table groups them by what had been observed when each was made. "Before data" means before any confirmatory verdict of the affected datasets had been read.

| When (UTC) | Change | Status when made |
|---|---|---|
| **Execution** | | |
| 09-23 07:45 | Request order randomized by trajectory rather than by request, so a shortfall leaves complete cells | Before data |
| 09-23 08:30 | Gateway failure; restart under a watchdog; lower concurrency | Before data |
| 09-23 10:28 | Execution moved to a standalone copy of the same code | Before data |
| 09-23 11:55 | Second gateway outage; failed requests re-sent in a fill-in pass | Before data |
| 09-23 12:15 | Worker reallocation to the mechanism queue once the primary run finishes | Before data |
| 09-23 13:13 | Data freeze moved from 09-26 to 09-27 00:00 because of throughput | Before data |
| 09-24 02:46–03:39 | Gateway admission limit raised to 16, then returned to 4 after upstream timeouts; primary run restarted with 12 workers | Before SWE-bench and mechanism data |
| 09-24 19:28–09-25 06:56 | Claude Opus 5.5 upstream outage; primary SWE-bench analysis restricted to three complete judges, Claude reported only on its matched partial sample; per-judge complete mechanism arms | SWE-bench and mechanism verdicts unread; τ-bench results known |

| When (UTC) | Change | Status when made |
|---|---|---|
| 09-25 | Primary H5 report-versus-control FPR tests corrected across three available judges; any partial Claude arm contrast reported descriptively | Before mechanism verdicts read; SWE-bench primary results known |
| 09-25 | H5 arm effect sizes gained task-bootstrap FPR/TPR intervals; any draws with no observations in a required label are counted as undefined | Before mechanism verdicts read; no randomized test or decision rule changed |
| 09-25 21:45 | Three available judges attempted all 18,900 three-arm requests; five Gemini upstream failures remained; Claude's 391 cached, report-eligible triples remain descriptive; data locked before the extended calendar cutoff after exhausting available requests | Before H5 verdicts read |
| **Design** | | |
| 09-23 11:16 | Mechanism experiment redesigned with a concurrent control and a placebo arm | Before data |
| 09-23 11:48–12:07 | Reference-adjudication sensitivity analysis and simulation study specified | Before SWE-bench data |
| **Analysis of AgentRewardBench (data public and already analyzed)** | | |
| 09-23 | Task identifiers normalized (implementation bug); missing released judgments excluded | After a first analysis |
| **Analysis changes registered before the SWE-bench and τ-bench data** | | |
| 09-23 11:48 | Label-specific H1 statistic; McNemar tests; within-task contrasts; task-mix/agent-specific split; conditional and marginal null references; basic bootstrap interval for H3; continuous version-pair estimates; transport anchored on all old-version tasks; τ-bench per-domain pairs and trajectory-level rates; audit intervals by size | Before data (first internal methods review) |
| 09-23 12:48 | Bootstrap of the pooled Youden index; task-cluster wild-bootstrap H1 test (primary for τ-bench); decomposition support; Proposition 3 diagnostics | Before data (second review) |
| 09-23 | Descriptive columns (outcome and error correlations, intervals for differences); disagreement-stratified audit comparator; guards for degenerate correlations and empty audit tables | Before data |
| 09-23 12:22, 12:48 | Hypotheses H6 (capability gradient) and H7 (solvability gradient) registered from AgentRewardBench patterns | Before data |
| 09-23 16:04 | Pipeline dry run on the SWE-bench development split; one bug fixed | After H6/H7 registration; confirmation data unread |
| **Exploratory analyses added after results** | | |
| 09-24 00:10 | τ-bench analyzed on completion, before SWE-bench data | τ-bench results known |

| When (UTC) | Change | Status when made |
|---|---|---|
| 09-24 02:30–04:30 | τ-bench rationale keywords and direct count of policy-violating turns; judge-aggregation variants; fixed-effects solvability slopes; wording corrections from a third review | τ-bench results known; SWE-bench unread |
| 09-25 | Signed pooled task-bootstrap intervals and 80%-power minimum detectable errors were added to the pair tables after the first three-judge SWE-bench analysis; the first review had called for these descriptions, but the initial implementation omitted them | Preliminary SWE-bench results known; no frozen test or decision rule changed |
| 09-25 | Result JSON writers made strict (`NaN` replaced by `null`); no estimate or test changed | Preliminary SWE-bench results known |
| 09-25 | A separate code audit found H3 bootstrap draws with undefined transport had been removed only from the numerator; all-or-nothing fixed-unit draws now give conditional intervals and undefined-draw rates; null references rerun under the same point-unit rule | Preliminary SWE-bench results known; H3 point ratio and registered decision unchanged |
| 09-25 | Nine initially incomplete issue IDs fixed from request-error metadata and removed for every agent and available judge as a whole-task missingness sensitivity; primary sample unchanged | Preliminary SWE-bench results known; fill-in not yet written |
| 09-25 | Reference-error sensitivity now corrects expected FPR *and* FNR using adjudicated errors from both reference-label classes; intermediate vote splits require interpolation | SWE-bench judge results known; adjudicator results unread |
| 09-25 | A task/label-conditioned permutation simulation added as a post-result robustness reference for H6 capability correlations | Preliminary SWE-bench H6 results known; no registered decision changed |
| 09-25 09:35 | Three-judge SWE-bench sample frozen at 8,743/8,750 cells after eight of 15 targeted Gemini retries succeeded; seven failures remain on three issues. Adjudication sample (489 cells) set by its fixed seed | Preliminary SWE-bench results known; no adjudicator response read |
| 09-25 | Corrected mistaken claim of public preregistration; repaired single-string `--cache-only` splatting in the finalizer | All outcomes known; no estimate or test changed |
| 09-25 | Cache-only replay exposed completion-order-dependent Monte Carlo draws and four net one-vote shifts in judge-by-label report-arm totals (two on failed patches) where some own-report/placebo prompts share a cache key; sorted analysis cells and made result merging preserve completed attempts and reject conflicts | All outcomes known; H5 decisions unchanged; final results regenerated without live calls |

Two further disclosures: automatic code formatting changed the bytes, but not the behavior, of two hashed files (the request cache keys, which hash the full prompts, match completions made before the change), and the analysis code itself had not been hashed at the freeze; it was hashed and pushed before the SWE-bench and τ-bench data were read.

### 4.7 Post-submission external-scaffold validation

The original manuscript and results above were complete when this additional design was fixed in a **private** repository on 2026-09-28 (commit `ce8c6a1`), before any external-cohort confirmation verdict was inspected. The original arXiv submission had not yet been publicly announced at that time. This is not the 2026-09-23 original preregistration, an independently public pre-outcome timestamp, or a design blind to the original H6 result. Two GPT-6-Sol-only smoke calls and one fixed development cell scored by each of the three available judges confirmed parseability and model IDs; their outputs did not choose the cohort or analyses.

The cohort was all eight specified OpenHands-labelled public submissions with verifiable prediction JSONL, execution `report.json` and actual graded evaluation-log `patch.diff` on at least 238 of the **same original 250 confirmation issues**. The single Agentless submission was fixed as a descriptive input-portability comparison, not a ninth member of the OpenHands capability analysis. The selection, graded artifacts, original task/sample/prompt hashes and source revision are documented in Appendix F. The three available original judges received the same patch-only prompt; none saw the execution label or agent name. First attempts used a new cache replicate, with at most one retry for an unparseable verdict. Upstream errors remained missing rather than being coded as rejections. The original unavailable Claude judge was not substituted. A further 100 original mini-SWE-agent cells (five agents, ten successes and ten failures each, 85 unique issues) were selected without consulting old verdicts and re-judged with a fresh replicate alongside the new cohort. These anchors monitor test-retest agreement, not independent truth.

Let C(a,j,h) be agent a's mean within-issue acceptance difference from other OpenHands agents sharing execution label h for judge j; h=0 gives the task-conditioned false-positive contrast, while C(a,j,1) - C(a,j,0) is its Youden contrast. The fixed external H1 permutes verdicts within (issue, execution-label) strata, with 9,999 draws and within-three-judge Holm adjustment. For external H6, each agent's test-based resolve rate is correlated with both contrasts using Spearman ranks; the three judge correlations are averaged. We permute the eight capability ranks *jointly* across judges and enumerate all 40,320 permutations, testing two-sided. The previously specified directional replication criterion requires a positive false-positive association and a negative Youden association, **both** with $p < 0.05$. Agent-level intervals use 2,000 task-cluster bootstrap draws. Prespecified checks exclude truncated patches, remove any initially source- or judge-incomplete issue across all eight agents, and omit each agent in turn; 200 further null draws shuffle verdicts within judge/issue/reference-label strata. These procedures concern error **relative to execution tests**; no new independent human correctness labels were collected.

## 5 Results

### 5.1 AgentRewardBench (replication, released evaluators)

The test split contains 1,106 expert-labeled trajectories of four web agents on 300 tasks; expert success rates range from 17.0% (Llama 3.3 70B) to 33.0% (Claude 3.7 Sonnet). On the 84 doubly annotated test trajectories, two experts agreed on 89.3% (Cohen's $\kappa = 0.78$), a reminder that the reference is itself noisy.

Only the H1–H4 statistics of the frozen plan were specified before these data were analyzed; the label-specific and cluster-robust tests, within-task contrasts, error decomposition, null references, McNemar tests, capability and solvability associations, and discordance comparisons in this section were added afterwards and are exploratory for AgentRewardBench.

**Differential error (H1).** Within-task permutation tests with the preregistered statistic reject non-differential error for 9 of the 15 evaluators at the unadjusted 5% level and for 5 after Holm adjustment within the dataset (AER-C, AER-V, GPT-4o with accessibility tree, and both Qwen2.5-VL judges); the registered decision adjusts across all judge-by-dataset tests and is reported in Section 5.5. The label-specific statistic rejects for 5 after Holm adjustment (AER-C, GPT-4o and both Qwen2.5-VL judges, and NNetNav); its cluster-robust version rejects for 4, excluding Qwen2.5-VL with screenshots (cluster-adjusted p = 0.070). The plan did not fix the set of "LLM judges": 5 of all 14 LLM-based evaluators reject, as do 3 of the 7 frontier or large open models used for the pilot's go/no-go rule. Neither meets the at-least-half criterion, including under the global adjustment in Section 5.5. The pre-canonical-order permutation run had counted 6 of 14, including a borderline Llama 3.3 result; its change after row-order normalization did not change the dataset-level decision (Section 4.6).

Task-conditioned contrasts show where the differential error lies (Figure 1). All 15 evaluators accept Llama 3.3's failed trajectories less often than the other agents' failed trajectories on the same tasks (mean −6.7 points, range −11.3 to −0.9; the interval excludes zero for 12), and all 15 accept Claude 3.7 Sonnet's failures more often (mean +5.0 points, range +0.3 to +11.9; 7 exclude zero). GPT-4o's trajectories are judged like everyone else's (1 of 30 contrasts excludes zero). The judges' false-acceptance rate thus rises with the agent's competence: the strongest agent's failures are the most convincing and the weakest agent's the least. The rank correlation between an agent's expert success rate and its within-task FPR contrast is positive for all 15 evaluators (mean 0.80), and that for the TPR contrast is negative for 12 (mean −0.43), so judges discriminate less well between the successes and failures of stronger agents (mean correlation with the Youden contrast −0.76). With four agents this pattern cannot reach significance (exact permutation p = 0.17); we registered it as hypothesis H6 for the 35 SWE-bench agents before reading their verdicts.

A second, task-level pattern points the other way. Judges accept a failure more often when other agents solved the same task: averaged over the 14 LLM-based evaluators, false acceptance rises from 12.4% on tasks no other agent solved to 23–26% on tasks that at least one other agent solved, and the slope is positive with a task-bootstrap interval excluding zero for all 15 evaluators. Part of this *solvability gradient* reflects composition: tasks come from different source benchmarks with different base rates, and agents differ in which tasks they fail. With agent and benchmark fixed effects the slopes roughly halve; they remain positive for all 15 evaluators but exclude zero for only 4. Failures on solvable tasks may be near misses that look like solutions, but the share of other agents that solved a task measures the task, not the focal trajectory, and conditioning on realized outcomes also selects tasks with more reference errors, so the association does not establish that interpretation. To the extent that the gradient holds, weaker agents, which fail more of the tasks others solve, receive higher marginal false-positive rates, while the agent-specific gradient favors stronger agents; marginal rates mix the two. We registered the solvability gradient (as the "near-miss gradient") as hypothesis H7 for SWE-bench. Marginal rates are accordingly noisy summaries: the GPT-4o judge with accessibility-tree input accepts between 4.7% and 17.9% of failed trajectories depending on the agent.

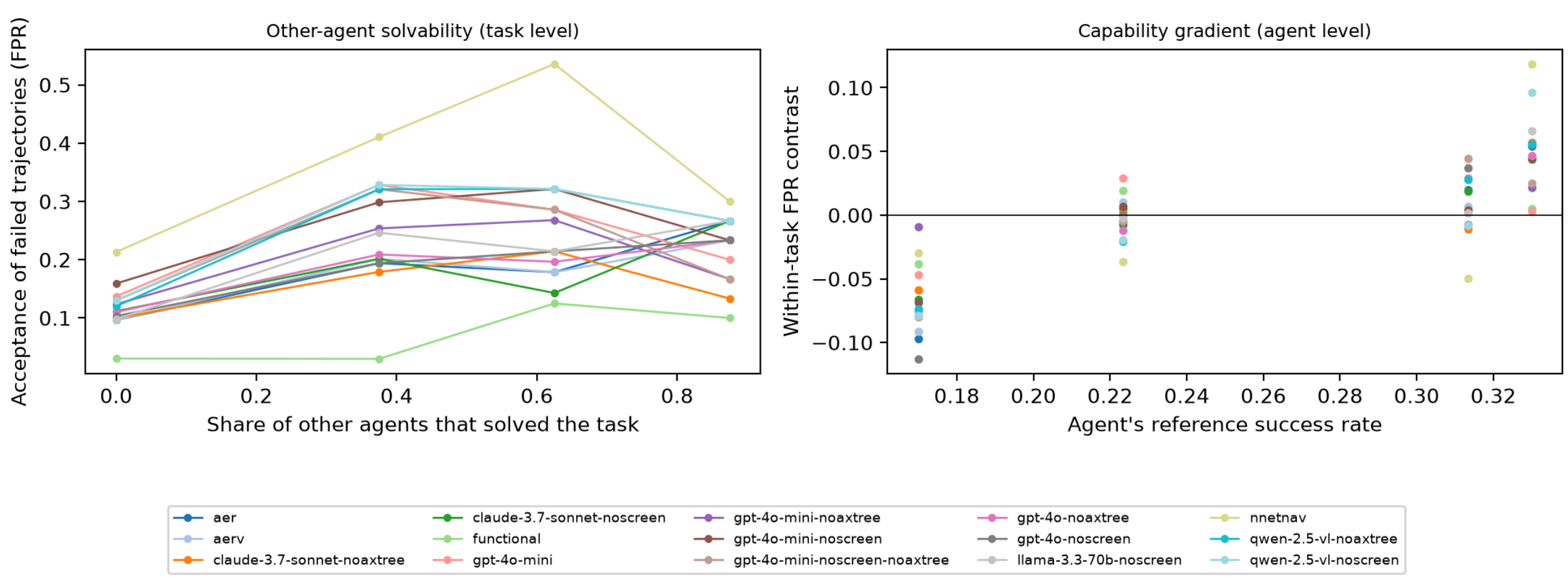

Figure 1. AgentRewardBench: false acceptance on failed trajectories by the share of other agents that solved a task (left), and within-task false-acceptance contrasts against each agent's expert-labeled success rate (right). The task-level and agent-level associations point in opposing directions on this dataset; neither identifies a causal effect.

**Agent comparisons (H2).** Across the 90 judge-by-agent-pair comparisons, the differential component of the comparison error is significant after false-discovery-rate control in 11.1% (10 of 90; the preregistered binomial test against 5% gives p = 0.015, but treats dependent comparisons as independent). This share exceeds its distribution under both simulated non-differential judges (conditional null mean 4.5%, 95th percentile 8.9%, p = 0.015; marginal null mean 1.6%, p = 0.005). The evidence is concentrated in one agent: 8 of the 10 significant comparisons involve Llama 3.3; without them 2 of 45 remain, and without the rule-based evaluator 8 of 84. The direction is consistent: relative to a non-differential judge of the same accuracy, judges widen the gap between Llama and stronger agents; for the five evaluators with a significant differential component for the Claude–Llama pair, it ranges from −7.5 to −12.2 points. Splitting the comparison errors, the agent-specific (leave-pair-out residual) component, with a mean absolute value of 4.0 points, is as large as the task-mix component (4.6 points); its interval excludes zero in 28.9% of comparisons, and its size exceeds the conditional null (mean 2.5 points, p = 0.005). Other agents' trajectories support the task-mix prediction for a median 86% of trajectories; the rest use pooled rates. Exact McNemar tests on tasks that both agents fail reject equal false acceptance in 26.7% of comparisons.

The consequences for decisions are mild on this dataset. Judge-only decisions disagree with reference-label decisions in 32.2% of comparisons, but this is *less* than a non-differential judge of the same accuracy would produce (conditional null mean 50.3%; marginal 46.0%), and every disagreement is a lost or spurious detection rather than a reversal: 20 comparisons in which the reference shows a significant improvement and 7 in which it shows a significant decline are judged inconclusive, and 2 without a significant reference difference are judged significantly worse. Point estimates reverse sign in 4 of 90 comparisons (4.4%, fewer than under either null): three GPT-4o-mini configurations rank GPT-4o above Claude 3.7 Sonnet, whose reference difference is 1.7 points, and the rule-based evaluator reverses the Qwen–GPT-4o comparison despite a 9-point reference difference. Because the differential error aligns with the true differences, judge-only tests retain more power than attenuation alone predicts: the median ratio of judge-only to reference test statistics is 0.70, against a median paired correlation $\rho_D$ of 0.46. The mean absolute comparison error is 4.3 points (maximum 13.1).

**Transported calibration (H3).** Calibrating each judge on one agent's labeled trajectories and transporting the Rogan–Gladen correction to another *increases* the mean absolute comparison error from 4.3 to 6.7 points: the ratio is 1.55 (preregistered percentile 95% CI 1.11–2.11; basic bootstrap CI 0.98–1.98), transport is worse than the uncorrected judge in 58.3% of the 180 ordered comparisons, and corrected rates were clipped in 2.2%. The preregistered criterion (lower bound above 0.5) is met, but, as Section 3.2 and Appendix C show, it would also be met by a judge without differential error. The informative comparison is with the simulated nulls: the observed ratio exceeds every simulated value under both the conditional null (mean 0.85, 95th percentile 0.97) and the marginal null (mean 1.00, 95th percentile 1.18). Transport fails here because the error rates it transports differ across agents, not merely because correction adds noise.

**Audits and efficiency (H4).** PPI++ intervals from paired audits cover the pool's reference difference in 89.8%, 93.3%, 94.1% and 94.5% of simulated audits of 20, 40, 80 and 160 tasks (91.5% to 94.7% with Student-t quantiles), whereas the judge-only interval covers it in 76.7% of comparisons (H4a supported for audits of at least 40 tasks). The judge buys little precision: at 80 tasks the mean interval width is 0.158 for the classical estimator that ignores the judge, 0.176 for untuned PPI and 0.138 for PPI++ (13% narrower than classical); oversampling the tasks on which the judge's two verdicts differ gives 0.146, with coverage of 93.3%. Dividing the judged difference by an audited Youden index is not a substitute: its root-mean-square error stays near 6 points as audits grow, whereas PPI++'s falls to 1.8 points at 160 tasks. The ideal efficiency gain is markedly smaller for paired differences than for levels: the median correlation between judged and reference outcomes is 0.62 for levels but 0.46 for paired differences, giving mean efficiency factors of 1.64 and 1.30 (one-sided Wilcoxon $p = 8.7 \times 10^{-17}$, descriptive; H4b supported); the median paired gain is 45% of the level gain. AgentRewardBench violates Proposition 3's non-differential assumption, so we use the proposition as a benchmark rather than a prediction. Its general plug-in form, $J^2 \, Var(D) / Var(\varepsilon_n - \varepsilon_o)$, predicts a mean paired factor of 1.42 (correlation with the observed factors 0.78); the observed 1.30 is lower. A descriptive pattern is compatible with Proposition 4: on the tasks on which the two agents' outcomes differ — the only tasks that carry information about the comparison — the failing agent's trajectory is accepted 21.8% of the time, against 13.0% for failures on tasks both agents fail (higher in 89% of judge-pair combinations), while successes are recognized at similar rates (76.7% and 80.1%). The label-matched Youden index is 0.55 on discordant tasks and 0.67 on concordant ones. Because discordant tasks are selected by the agents' realized outcomes, and are enriched for reference errors and for the solvability gradient above, this does not identify task-level discriminability, and the judges also show agent-specific error here, so Proposition 4's assumption does not hold exactly. Judge errors on the two agents' trajectories of a task are positively correlated (median 0.22 over all tasks), which recovers part of the loss.

**Exploratory: same-family judging.** Evidence of self-preference is mixed. The Claude 3.7 judge with screenshots accepts Claude's failed trajectories 5.7 points more often than other agents' failed trajectories on the same tasks (95% CI 0.6–10.7), and the Qwen2.5-VL judge with screenshots accepts Qwen's successful trajectories 10.8 points more often (CI 2.0–20.2), whereas the Llama judge accepts Llama's failures 7.9 points *less* often (CI −13.4 to −2.5), in line with every other judge. The remaining same-family contrasts include zero. Given 14 same-family contrasts, three nominally significant results are compatible with a modest or absent self-preference effect.

### 5.2 SWE-bench Verified (primary)

The 35 mini-SWE-agent submissions produced a patch and an execution-based resolution label for each of 250 held-out issues (8,750 agent-task cells). After the Claude Opus 5.5 upstream outage, the modified primary analysis has aligned analysis rows from all three available judges on 8,743 cells, including one unparseable verdict coded negative under the frozen rule: 99.9% of the intended three-judge pool, spanning all 35 agents and 250 tasks. Eight of the 15 initial Gemini failures were recovered from the unchanged prompts after a targeted retry pass; seven requests on three issues exhausted the ten-attempt limit and remain missing. The primary sample is frozen at this point. Across the common cells, agents' test-based resolution rates range from 20.0% to 77.6%. Of the 20 preregistered version pairs, nine have a positive reference difference whose 95% paired interval excludes zero, eleven are inconclusive, and none has a significant decline. The Claude judge is analyzed only on its 7,648 four-judge-matched cells (87.4% of the planned pool), not imputed or replaced by a second GPT-6 Sol verdict.

**Differential judge error (H1).** Every available judge rejects the within-(task, reference label) exchangeability null with both the preregistered and label-specific statistics (9,999 permutations, p = 0.0001 per judge; within-dataset Holm p = 0.0003). The original rule required four complete judges, so it cannot literally be met; the three-judge result is a clearly marked outage modification. The effect is not just statistically detectable. Across the 35 agents, the median false-positive rates are 65.7% for Gemini 3.5 Flash, 66.0% for GPT-5 mini, and 40.0% for GPT-6 Sol; their respective ranges are 7.5–95.0%, 9.5–90.0%, and 2.0–69.0%. Uncertainty for agents with few failed patches is larger than for the others (per-agent Wilson intervals are in the released aggregate tables). Median false-negative rates are 9.6%, 13.1%, and 22.2%. The judge can thus be highly accurate on successful patches yet accept many incorrect nonempty patches.

**Agent capability and task solvability (H6 and H7).** The registered, task-conditioned H6 pattern from AgentRewardBench replicates strongly in this different domain. Across 35 agents, the Spearman correlation of reference resolve rate with within-task false-positive contrast is 0.827 (Gemini), 0.764 (GPT-5 mini), and 0.865 (GPT-6 Sol); the mean is 0.819 and the permutation p-value is 0.0001. Successful-patch acceptance also rises with capability, but false acceptance rises faster: correlations with the task-conditioned Youden contrast are −0.783, −0.718, and −0.741 (mean −0.747, p = 0.0001). The top half of agents has a task-conditioned false-positive contrast 21–26 points higher than the bottom half. This is an association among public configurations of one scaffold, not evidence that raising an agent's capability causes the judge to change its behavior. In a post-result check that shuffles verdicts only within each (judge, task, reference label) stratum, holding outcomes and capabilities fixed, the mean correlations are −0.001 for FPR and +0.004 for Youden; none of 200 simulated values is as extreme as +0.819 or −0.747 (two-sided Monte Carlo p = 0.005 for each). Thus the H6 pattern is not an automatic result of agents solving different tasks under a task-conditionally non-differential judge.

H7, however, fails in the **opposite direction** from the positive solvability gradient registered from AgentRewardBench. Among failed patches, increasing the share of *other* agents that solved the same issue from zero to one is associated with a **decrease** of 48, 38, and 39 points in false acceptance by Gemini, GPT-5 mini, and GPT-6 Sol. All three 95% task-bootstrap intervals exclude zero on the negative side (Gemini −58 to −39 points; GPT-5 mini −48 to −28; GPT-6 Sol −51 to −29). After agent and repository fixed effects the slopes are smaller (−24, −20, and −21 points), but their intervals still exclude zero. Excluding empty patches does not change their signs; neither does it remove H6 (mean correlations +0.777 for FPR and −0.694 for Youden). The task-level pattern is therefore *not* a universal near-miss effect: an issue that many agents solve may make a failed code patch easier, not harder, to reject. Task difficulty, patch characteristics and possible reference error could also contribute, so this association is descriptive. Figure 2 contrasts the task and agent gradients directly.

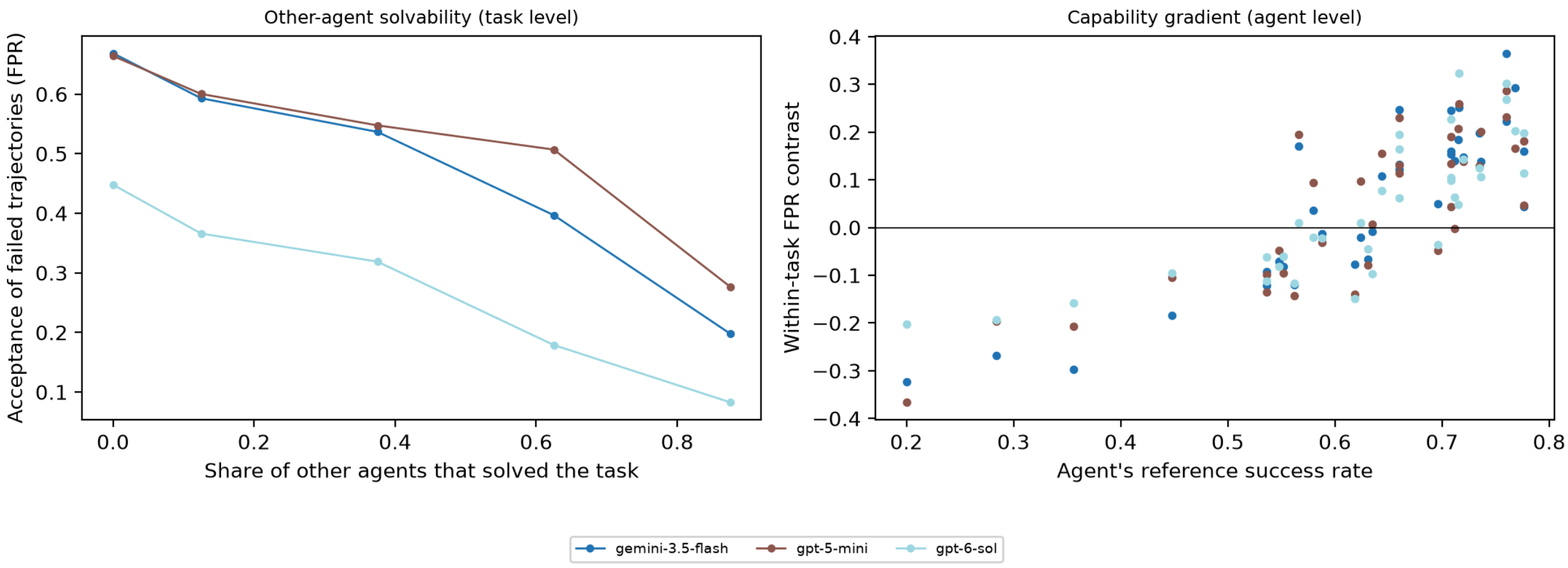

Figure 2. SWE-bench: false acceptance of a failed patch falls as more other agents solve its issue (left), but each judge's within-task false-acceptance contrast rises with the agent's reference resolve rate (right). Each point on the right is an agent-judge combination; the registered H7 prediction was positive and is contradicted.

**Version comparisons (H2).** Table 1 separates agreement on the broad ranking from errors in release decisions. All three judges have a positive Kendall correlation with the reference ordering of 35 agents (0.71–0.79). Nevertheless, the differential comparison component e_d = D_J − J_pool D_H has a two-sided bootstrap sign-test p-value below 0.05 after within-judge Benjamini–Hochberg adjustment for 32 of the 60 judge-by-version-pair units (53.3%). Separately, 34 of 60 *unadjusted marginal* 95% bootstrap intervals exclude zero; these are not FDR-adjusted confidence intervals. Only about 1.1% of units meet the same criterion under a simulated conditionally non-differential judge that preserves task-level judging difficulty (95th percentile 5.0%; Monte Carlo p = 0.005). The dependence-ignoring binomial test was specified in the frozen plan but is not the basis of this inference. Averaging over the 60 prespecified units in their old-to-new direction, the signed comparison error is +2.8 points (joint task-bootstrap 95% interval +1.6 to +4.2), and the differential component is +6.8 points (+5.8 to +7.7). The mean absolute comparison error is 3.8 points, up to 11.6 in one unit. The leave-pair-out agent-specific residual averages 6.4 absolute points, compared with 3.5 points for the task-mix component; other agents provide task/label support for a median 99.8% of the focal trajectories.

| Judge | Median FPR / FNR | Nonzero e_d after FDR | Release-decision disagreements | Mean absolute e | Mean absolute transported error |
|---|---|---|---|---|---|
| Gemini 3.5 Flash | 65.7% / 9.6% | 12 / 20 | 4 / 20 | 3.3 points | 20.3 points |
| GPT-5 mini | 66.0% / 13.1% | 9 / 20 | 4 / 20 | 2.9 points | 21.8 points |
| GPT-6 Sol | 40.0% / 22.2% | 11 / 20 | 3 / 20 | 5.1 points | 16.5 points |

*Table 1: Three-judge SWE-bench version comparisons on common tasks. FPR/FNR are medians across 35 agents; e and e_d are defined in Section 3. Units are judge-by-pair, not independent experiments; the last column uses the previous version's error rates to correct the next version's judged score.*

The discrete decisions disagree in 11 of 60 units (18.3%): eight judge-only intervals declare an improvement where the reference interval is inconclusive, and three miss a reference-detected improvement. There is no *significant* better-to-worse reversal among these 20 pairs, although 11 of 60 judged point differences reverse the reference sign. This reversal rate does not exceed its conditionally non-differential simulation (mean 21.2% versus observed 18.3%). Neither does the decision-disagreement rate, which the simulated null puts near 32.7%. These are **judge-only significant,** **reference-inconclusive** results, not proof that the eight claimed upgrades are actually harmful or null: six of their reference point estimates are positive. The practical risk is acting with greater confidence than the reference supports, not a large number of confident reversals. For example, a combined scaffold and configuration change from Claude Sonnet 4.5 to its high-effort successor changes the test-based solve rate by −0.4 points (95% interval −5.2 to +4.4), yet GPT-5 mini and GPT-6 Sol judge it +8.0 points (+2.0 to +14.4) and +9.6 (+3.2 to +16.0), respectively. All three judges declare the Claude Opus 4 to 4.5 change an improvement of 10–16 points, although its reference improvement is 4.0 points (interval −1.2 to +8.9). The source of excess confidence matters: for a *nominal, unadjusted* two-sided 5% test of e_d, the approximate median 80%-power detectable magnitude is 8.4 points (range 4.6–10.1); adjustment across 20 pairs requires still more power. Half of the reference version differences are smaller than five points. Continuous estimates and intervals, not just a count of discoveries, are therefore essential (Appendix B and Figure 3).

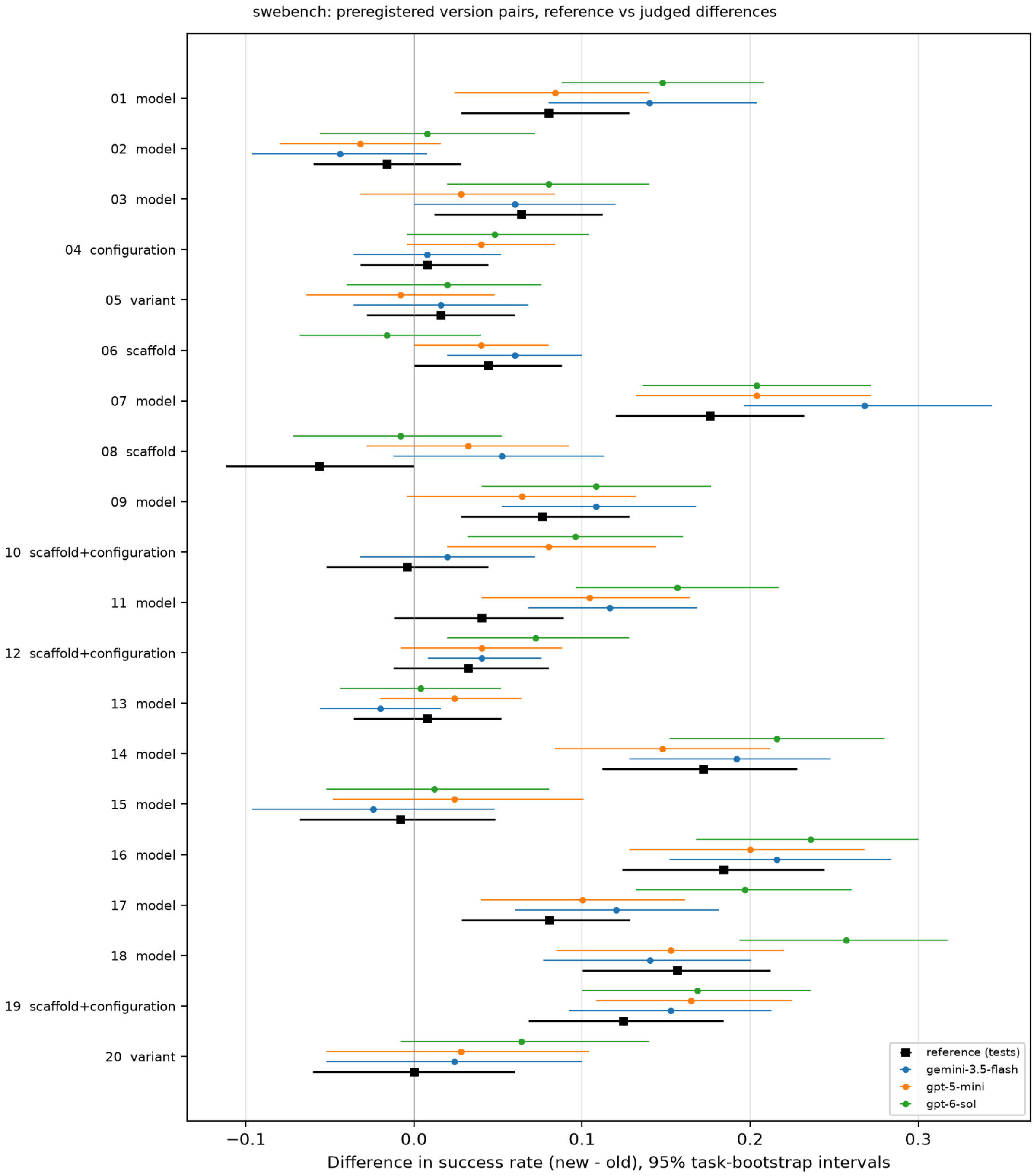

Figure 3. Reference and judged paired differences for the 20 prespecified SWE-bench version pairs. Pair numbers and types on the vertical axis correspond to Appendix B; error bars are *marginal*, not simultaneous, 95% intervals computed from jointly resampled tasks. A judge-only interval can exclude zero where the reference interval does not.

**Transported calibration (H3).** Applying the old version's estimated sensitivity and false-positive rate to the new version is worse than the uncorrected comparison in **59 of 60** ordered judge-pair units. Mean absolute error rises from 3.8 to 19.5 points: R = 5.20 (percentile bootstrap interval 3.46–5.19; basic interval 5.20–6.93). The intervals are conditional on all point-defined transports remaining defined: 491 of 2,000 joint task-bootstrap draws (24.6%) made at least one old-version Youden index nonpositive and were omitted **in full**. They are not unconditional 95% coverage guarantees near this boundary. The corrected rate had to be clipped to [0, 1] in 35.0% of units; restricting attention to unclipped units still yields an error ratio of 4.98. The preregistered lower-bound-above-0.5 decision is met but, as Section 3 and Appendix C show, not informative by itself. The conditional non-differential simulation gives a mean ratio of 1.71 (95th percentile 2.11), and the marginal simulation 1.47 (95th percentile 1.74); all 200 simulated point ratios under each null were defined, and the observed ratio exceeds every one ($p = 0.005$). This null comparison is more informative than the conditional bootstrap interval for whether differential error contributes. Judge-specific median Youden indices on these version pairs are only 0.19–0.38. Calibration divides version-dependent error *and* sampling noise by that index, explaining why transport is particularly unreliable for close comparisons.

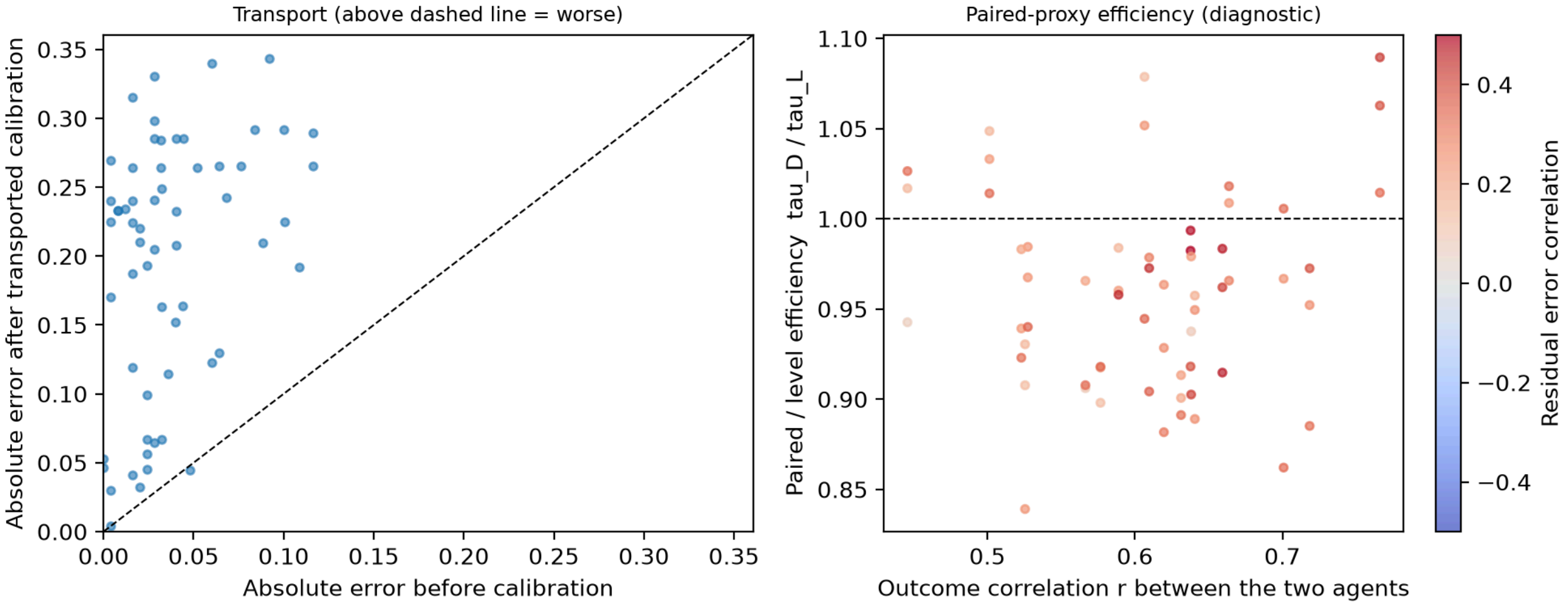

Figure 4. On the prespecified SWE-bench version pairs, transporting old-version calibration raises absolute comparison error in 59 of 60 judge-pair units (left; above the diagonal is worse). The ratio of paired to level label efficiency is usually below one (right); color denotes the correlation of the judges' residual errors on paired tasks.

**Audits and efficiency (H4).** At 80 randomly selected task pairs, a classical interval using only execution labels has mean width 14.9 points and covers the pool's difference in 94.3% of simulated audits. An untuned PPI interval is *wider*, at 19.8 points; PPI++ narrows the width to 14.2 points (about 5%) with 94.1% normal-quantile coverage (94.5% with Student-t quantiles). PPI++ coverage rises from 89.5% at 20 audited tasks to 94.4% at 160; these are *empirical*, not exact finite-sample, coverage statements. Judge-only intervals cover the reference difference in 80.0% of units. The ideal paired-difference efficiency factor is 1.084 compared with 1.133 for the mean of the two level factors; the median gain above one is 46.6% of the level gain. The median paired correlation is 0.213 versus 0.314 for levels, while the two agents' reference outcomes correlate at 0.615 across tasks. The general plug-in formula of Proposition 3 gives 1.080 for the paired factor, close to the observed 1.084, but the judges violate its non-differential assumption, so this match is a diagnostic rather than a validation. Disagreement-stratified auditing has width 15.0 points at 80 tasks, showing that spending more labels on judge-discordant tasks does not automatically beat a tuned, unbiased paired estimate.

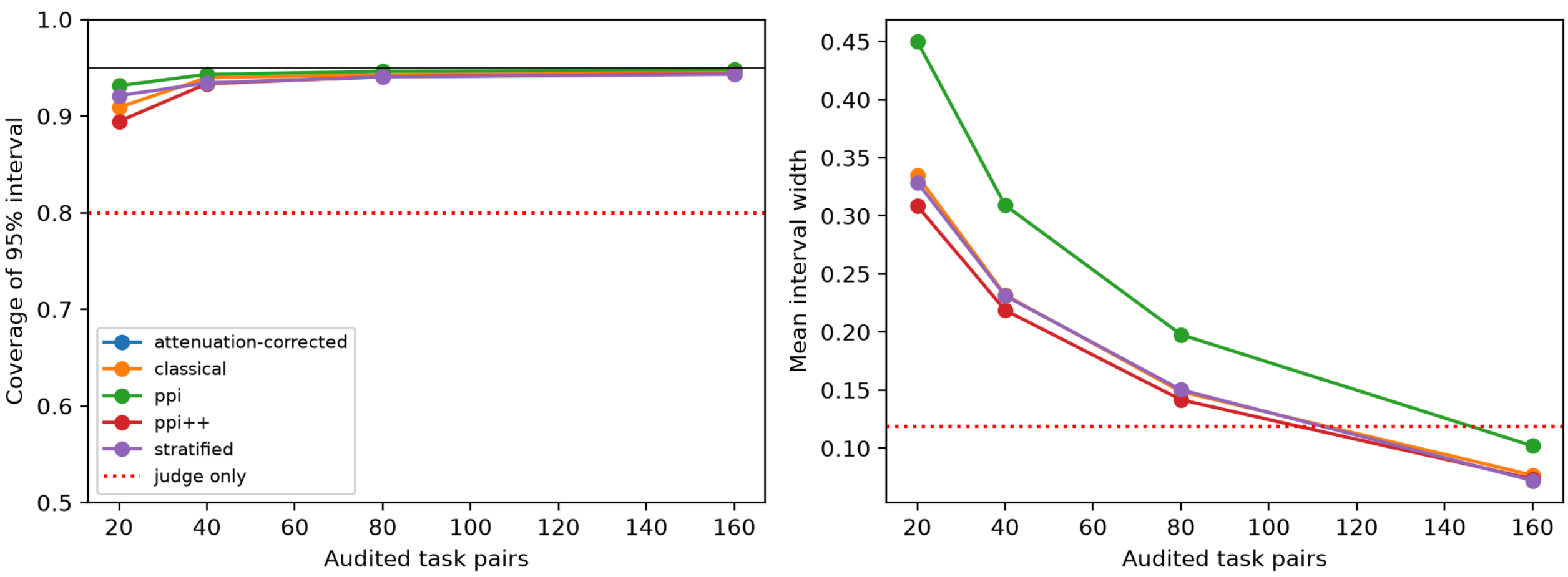

Figure 5. Simulated 95% interval coverage and mean width by audit size for the SWE-bench version pairs. PPI++ is modestly narrower than the classical paired audit, while untuned PPI is wider; the horizontal red line represents coverage or width of the judge-only interval.

**Sensitivity and aggregation.** Excluding the 271 empty-patch cells (all unresolved and rejected by each judge) and one additional invalid-verdict cell leaves 8,471 paired cells and does not remove the pattern: H1 rejects for all three judges, the fraction of nonzero e_d units is 48.3% instead of 53.3%, and the transported-error ratio is 5.34 instead of 5.20. About 52% of this sensitivity's joint bootstrap draws leave some transported corrections undefined. The release-decision disagreement rate does change, from 11/60 to 13/60, demonstrating that counts near a significance threshold depend on the specified input population. The H6 FPR and Youden correlations remain +0.777 and −0.694, while all three H7 slopes stay negative. On the 7,648 cells judged successfully by all four prespecified models, Claude Opus 5.5 also rejects H1 and has seven of 20 nonzero e_d components; its mean absolute comparison error is 3.5 points and its transported error 5.2 points, much less extreme than the other judges' transport failures, although it disagrees with the reference decision for 6 of 20 pairs (the three other judges disagree for 5, 3 and 3 of 20 on these matched cells). The four-judge matched-subset transport ratio is 4.04, not 5.20; we do not use this selected subset to replace the three-judge primary result. Finally, a strict majority of the three available judges still rejects H1 and has 11 of 20 nonzero e_d components, with three release-decision disagreements and 4.3-point mean absolute error. Averaging their probabilities lowers absolute error to 3.0 points but does not remove differential error or decision discrepancies. As an exploratory label-efficiency check, averaging the three binary verdicts raises the *ideal* paired efficiency factor from 1.084 for the average single judge to 1.145; averaging their reported probabilities gives 1.111. Individual probabilities do not consistently beat binary verdicts, and three calls cost more than one. Proposition 3 describes a single binary proxy, not a limit on richer proxies; even this ensemble's gain is modest. Unlike on τ-bench, ensembling is not a general fix.

Because the initial 15 Gemini request failures were concentrated on nine issues, we ran a *post-result* whole-task sensitivity that excludes each such issue for every agent and judge (241 tasks, 8,435 aligned agent-task cells). The resulting H1 tests still reject for all three judges; 27 of 60 e_d tests reject after within-judge adjustment, versus 32 of 60 in the primary sample; decision disagreements are 10 of 60 versus 11 of 60 in the primary sample, and the transport error ratio is 5.08. The mean capability correlations stay +0.815 for FPR and −0.745 for Youden, and all three solvability slopes remain negative. This does not establish that the failed requests were missing at random, but it shows that their task-level concentration does not explain the main patterns on the observed tasks.

**Reference-error sensitivity.** A blinded, separate GPT-5.5 adjudicator received the issue, maintainer patch, test changes and candidate patch but neither the execution label nor the primary judges' verdicts. It returned a parseable verdict for all 489 stratified agent-task cells (35 agents; Section 4.5). It disputes the execution label for 41 of 175 failed patches accepted by at least two primary judges (23.4%, Wilson interval 17.8–30.2%) and 86 of 175 test-resolved patches rejected by a majority (49.1%, interval 41.8–56.5%). Among the unanimous judge–test agreements the dispute rate is 2 of 69 failures (2.9%) and 2 of 70 successes (2.9%). These rates are *conditional on vote/label stratum* and cannot be averaged as if the sample were uniform over all 8,743 cells. Of the 41 adjudicator disagreements among judge-accepted test failures, 18 (43.9%) identify tests whose requirements may be more specific than the issue, according to the adjudicator's separate `tests_overspecified` tag.

If stronger agents' test failures were disproportionately mislabeled, their apparent judge false positives could be partly an artifact of the reference. We find no such gradient in this adjudicator: among the 175 judge-accepted test failures, the agent-level Spearman association between its dispute rate and test-based resolve rate is −0.003; permuting agent capabilities as clusters gives $p = 0.950$ for the cell-level slope. The corresponding association for majority-rejected test successes is −0.256 (descriptive Spearman $p = 0.138$; agent-cluster slope $p = 0.135$).

As a more assumption-dependent calculation, we reweight each agent's FPR and FNR by the adjudicator's shrunken dispute rate for its outcome/vote stratum (including interpolation in the two unsampled intermediate-vote strata). Across agents, mean FPR falls from 58.9% to 56.7% for Gemini, 60.6% to 56.6% for GPT-5 mini, and 36.7% to 34.3% for GPT-6 Sol. The Spearman association of FPR with agent resolve rate is nearly unchanged: 0.890 to 0.889, 0.848 to 0.814, and 0.899 to 0.888, respectively. Expected FNR falls as well. These numbers neither correct the *task-conditioned* H6 estimand directly nor prove the execution labels wrong: the adjudicator sees the maintainer solution and may share biases with the judges. In particular, a model's critique of an execution-resolved patch is not a replacement for running stronger tests or having an independent engineer verify the issue's requirements. The registered reference-error sensitivity therefore does not explain away the observed capability/error association, but cannot exclude agent-dependent reference error that this adjudicator fails to detect.

### 5.3 τ-bench (conversational agents, four trials per task)

The confirmation set contains 155 tasks (47 airline, 108 retail), each attempted four times by Claude 3.5 Sonnet (new) and by GPT-4o, so every judge assessed 1,240 conversations; all 4,960 verdicts are valid (one after re-sampling). By the environment reward, Claude succeeds on 44.7% of airline and 71.1% of retail attempts, GPT-4o on 42.6% and 62.0%. Judging these conversations is hard, and the four judges operate at very different points: averaged over the two agents, GPT-5 mini accepts 57% of failed conversations (false-negative rate 19%), GPT-6 Sol rejects 79% of successful ones (false-positive rate 7%), and Claude Opus 5.5 (28% and 25%) and Gemini 3.5 Flash (34% and 42%) lie in between.

**Differential error (H1).** All four judges reject non-differential error with every statistic, including the cluster-robust test that resamples whole tasks ($p \le 0.0015$ after Holm adjustment); H1 is supported on τ-bench. The within-task contrasts show two opposite distortions. Claude Opus 5.5, Gemini 3.5 Flash and GPT-5 mini accept Claude's failed conversations far more often than GPT-4o's failed conversations on the same tasks (by 22.8, 16.6 and 19.3 points; all intervals exclude zero), favoring the stronger agent. GPT-6 Sol instead recognizes Claude's *successful* conversations far less often than GPT-4o's (by 22.8 points) and accepts GPT-4o's failures slightly more often (by 4.7 points), favoring the weaker agent. With two agents, capability cannot be separated from agent identity or model family, so this pattern resembles H6 but cannot test it.

**Comparisons (H2).** The plan compares the two agents within each domain, giving eight judge-by-domain comparisons; the differential component is significant in six (two of four in airline, four of four in retail). In airline, where the reference difference is small and inconclusive (GPT-4o 2.1 points lower; 95% CI −11.7 to 8.0), Claude Opus 5.5 declares GPT-4o significantly worse by 12.2 points (CI −23.4 to −1.6). In retail, the reference shows GPT-4o 9.0 points worse (CI −16.0 to −2.5). Three judges reach the same decision but exaggerate the gap (11.3 to 21.1 points), and GPT-6 Sol reverses it with confidence: by its verdicts GPT-4o is 17.8 points *better* (CI 11.8 to 24.1). A developer who used this frontier judge to choose between the two agents would have shipped the worse one with an interval that excludes the truth by a wide margin.

**Transport (H3).** Rogan–Gladen transport from one agent to the other is worse than the uncorrected judge in all 16 ordered comparisons; the ratio of mean absolute errors is 4.0 in airline and 4.1 in retail, with transported errors averaging 28 and 46 points, because the judges' Youden indices are small (GPT-6 Sol's pooled index is 0.11 in retail, GPT-5 mini's 0.13) and differences in error rates are divided by them. Corrected rates fall outside [0, 1] in 75% of retail comparisons. The conditional bootstrap intervals for R are 2.09–5.44 in airline and 2.80–4.69 in retail; 1.7% and 24.6% of joint bootstrap draws, respectively, are undefined because a sampled old-version Youden index is nonpositive. A single-trial non-differential simulation reference is not justified for these correlated trials, so the ratios and conditional intervals are descriptive rather than a valid test of version-specific judge error.

**Audits and efficiency (H4).** Paired audits remain valid (PPI++ coverage 92.1%–94.4% for 20 to 80 audited tasks), whereas judge-only intervals cover the reference difference in 75% of airline and 50% of retail comparisons. But the judges carry almost no information about the comparison: the median correlation between judged and reference paired differences is 0.19 in airline and 0.14 in retail (0.43 and 0.25 for levels), so the paired efficiency factors are 1.04 and 1.07 and PPI++ intervals are barely narrower than classical ones (0.167 versus 0.173 at 40 retail tasks). Over all 155 tasks, the best single judge, Claude Opus 5.5, reaches a paired correlation of 0.34; Gemini's is zero.

**Why the frontier judge reversed the comparison (exploratory, after the τ-bench results).** Both the conversations and the judge's rationales point to one mechanism. The τ-bench policies state that an agent should not call a tool and respond to the user in the same turn. Counting directly in the conversations, every one of Claude's conversations contains such turns — 7.4 per successful retail conversation on average — whereas 43% of GPT-4o's successful retail conversations contain any, with 0.5 per conversation on average (`scripts/mixed_turns.py`). The environment reward checks the final database state and ignores this rule. GPT-6 Sol rejects 350 of Claude's successful conversations, and 31.7% of those rationales cite the rule, against none of its 236 rejections of GPT-4o's successful conversations (keyword classification of the rationales, `scripts/rejection_reasons.py`; the rationales state the judge's reasons but do not by themselves verify each violation). Our judge prompt, like most, asks whether the agent "followed the policy", so a strict judge measures a stricter construct than the reference, and a difference in the agents' habits becomes a difference in judge error. Calibrating the judge on the evaluated agent's own labeled conversations could correct its aggregate rate, but calibration transported from the other agent cannot, because the habits — and therefore the error rates — differ between the agents.

**Exploratory.** Aggregating the four judges helps here, but how much depends on the aggregation rule. Two-to-two splits occur on 27% of conversations (31% of Claude's, 23% of GPT-4o's). Counting ties as rejections (three of four must accept), the aggregate shows no detectable differential error (cluster-robust $p = 0.16$) and comparison errors of 2.7 points in airline and 0.5 in retail; averaging the judges' stated probabilities gives errors of 1.1 and 0.7 points but detectable differential error ($p = 0.006$);

counting ties as acceptances exaggerates the retail gap by 8.3 points ($p < 0.001$). All three aggregates reach the reference decision in both domains. The judges' distortions point in opposite directions and partly cancel; Section 5.2 asks what happens when they point the same way. Same-family effects are mixed and confounded with capability with only two agents: Claude Opus 5.5 favors Claude's failures and GPT-6 Sol favors GPT-4o, but GPT-5 mini, also an OpenAI model, disfavors GPT-4o. Acceptance of failures does not change detectably with the other agent's success on the task (no near-miss gradient; the reference for this is a single other agent).

### 5.4 Randomized self-report mechanism

The patch-only primary condition deliberately hides the agent's final natural-language message. To isolate this input channel, we randomized three prompt arms within each of 2,100 (agent, issue) cells from 21 agent configurations and 100 held-out issues: a fresh patch-only control, the patch with the agent's *own* final message, and a placebo with the same agent's final message from another issue. Each arm was scored once by each available judge, with arms randomly ordered and adjacent within a cell. The three-model run attempted all 18,900 planned requests. Gemini had five persistent upstream errors; after the frozen one-resample parsing rule, there are 2,088 fully parseable, report-eligible triples for Gemini, 2,090 for GPT-5 mini, and 2,091 for GPT-6 Sol. Under the frozen primary rule, the one GPT-5-mini report-eligible cell with a persistently unparseable verdict is coded negative rather than omitted, so its primary analysis includes 2,091 cells, 821 of them failed patches. The partial Claude cache provides 391 comparable three-arm cells (12–25 per agent), reported descriptively rather than placed in the primary three-judge multiplicity family.

**Primary false-acceptance test (H5).** The predicted increase in false acceptance is **not observed**. On test-unresolved patches, adding the agent's own final message instead of the concurrent patch-only control changes the acceptance rate by −1.47 percentage points for Gemini (95% task-bootstrap interval −4.16 to +0.99; 819 cells), −0.49 for GPT-5 mini (−2.84 to +1.67; 821 cells), and +0.61 for GPT-6 Sol (−1.83 to +3.51; 821 cells). The one-sided task-cluster sign-flip p-values are 0.884, 0.700 and 0.365, respectively; Holm-adjusted across the three available judges, **all are 1.0**. Two of the point estimates have the opposite sign from H5. The strongest positive effect compatible with each individual 95% interval is roughly 1–3.5 points, so the experiment provides evidence against large *average* false-acceptance effects under this patch-only prompt, although marginal bootstrap intervals do not exclude every positive effect, and it cannot rule out smaller effects or effects of other ways of showing a full trajectory.

**Content, placebo and discriminability.** Replacing an unrelated same-agent message with the patch's own message changes false acceptance by −1.22, +0.24 and +0.73 points for Gemini, GPT-5 mini and GPT-6 Sol (two-sided $p$ = 0.279, 0.923 and 0.491). The placebo-minus-control false-acceptance changes are −0.24, −0.73 and −0.12 points; none is distinguishable from zero. On resolved patches, the report-minus-control true-positive changes are −0.47, −0.39 and −0.39 points, all with intervals including zero. Changes in Youden's J, the judge's separation of successful and failed patches, are +0.99 (interval −1.75 to +3.82), +0.09 (−2.61 to +2.67) and −1.00 points (−3.92 to +1.86). A partial Claude analysis also gives an inconclusive false-acceptance change of +0.69 points (−4.26 to +5.48; 145 failed-patch cells). In 17 of 2,100 planned cells, the own and placebo final messages produce identical prompts, so the content contrast has no information there. A cache replay later shifted four judge-by-label report totals by one net vote each; shared prompt keys are a plausible source, but the original row identities were overwritten (Section 4.6). The pooled contrasts provide no evidence that the added text explains the agent-dependent patch-only errors of Section 5.2. The null average effect does not exclude agent-specific effects in either direction; the heterogeneity analysis below is exploratory.

**Heterogeneity and repeatability.** A task-stratified permutation test finds nominal heterogeneity across agents for Gemini's report-versus-control false-acceptance effects ($p = 0.032$; agent means range from −20 to +11 points), but not for GPT-5 mini ($p = 0.452$) or GPT-6 Sol ($p = 0.866$). This is exploratory: there are multiple arm, label and judge contrasts, and a range over 21 small agent strata is especially noisy. Re-scoring the patch-only input in a fresh control request, much later than its original primary request, yields agreement of 92.4%, 88.5% and 90.5% for Gemini, GPT-5 mini and GPT-6 Sol (Cohen's $\kappa = 0.80$, 0.69 and 0.80). This is not a pure measure of stochastic decoding: it includes time-dependent upstream behavior. Excluding empty patches and invalid verdicts does not change the primary conclusion: all three adjusted p-values remain 1.0, and the largest report-minus-control false-acceptance point effect is only +0.40 points (GPT-6 Sol). H5 therefore fails its prespecified directional test on the available judges; a claim of self-report contamination from this intervention would not be supported.

### 5.5 Cross-benchmark synthesis

Table 2 collects the preregistered comparisons without treating fundamentally different tasks, judges or reference standards as exchangeable replicates. H1 is adjusted jointly over the 22 available judge-by-dataset tests: five of 14 LLM-based AgentRewardBench evaluators reject (short of its half-of-judges criterion), three of three available SWE-bench judges reject (the originally planned fourth was unavailable), and all four τ-bench judges reject. AgentRewardBench shows a task-conditioned pattern but does not meet the half-of-judges H1 criterion under either reported definition of its judge set. The other two datasets meet the available-judge criterion, subject to the explicitly logged change in SWE-bench coverage.

| Dataset and reference | H1 (global Holm) | H2: nonzero e_d / judged comparisons | Disagreeing release decisions | H3: transport / naive absolute error | Ideal paired / level label-efficiency factor |
|---|---|---|---|---|---|
| AgentRewardBench, expert labels | 5 / 14 LLM evaluators | 10 / 90 | 29 / 90 (32.2%) | 1.55 | 1.30 / 1.64 |
| SWE-bench Verified, execution tests (three available judges) | 3 / 3 | 32 / 60 | 11 / 60 (18.3%) | 5.20 | 1.08 / 1.13 |
| τ-bench airline, database reward | 4 / 4 across both domains | 2 / 4 | 1 / 4 | 4.03 | 1.04 / 1.30 |
| τ-bench retail, database reward | same four | 4 / 4 | 1 / 4, including a confident reversal | 4.14 | 1.07 / 1.16 |

*Table 2: H1 counts refer to a family of 22 judge-by-dataset tests, not separate significance tests on independent agent pairs. H2 units share agents and tasks. For τ-bench, H1 is the task-cluster-robust test, and the three-judge SWE-bench result is a post-outage modification. Comparisons among the rows are descriptive, not a cross-benchmark meta-analysis.*

Two regularities survive the changes in domain and reference. First, a *fixed* judge can have error that varies with the evaluated agent, but the consequences differ: AgentRewardBench mostly loses power, SWE-bench has eight judge-only significant upgrades for reference-inconclusive pairs, and τ-bench includes one confident choice of the worse agent. The pooled mean comparison error detects systematic bias but can hide errors of opposite sign on individual pairs; deciding one release requires that pair's reference audit and interval, not a leaderboard rank correlation or a test of pooled bias.

Second, transporting error rates from a previously labeled agent can be worse than using the raw judge: all three datasets have a ratio above one, although the size varies sharply by judge, reference standard and Youden index. Only AgentRewardBench and SWE-bench have valid single-trial simulation references for the transport ratio; the τ-bench ratio is descriptive.

The more interesting contrast is what *does not* replicate. H6's agent-level gradient is strong among 35 SWE-bench configurations and directionally consistent with AgentRewardBench's four agents and three of four τ-bench judges, but a two-agent domain cannot identify a general capability gradient. H7's task-level prediction reverses sign from AgentRewardBench to SWE-bench, even after agent and task-group adjustment. There is no universal rule that a failure on a task that others solved is harder to judge. These are conditional associations, not identified mechanisms; the randomized self-report intervention (Section 5.4) adds a useful negative result. Giving the judge the agent's own final message did not systematically raise false acceptance above a fresh concurrent control, and neither an unrelated same-agent message nor a test-retest resample establishes a universal mechanism for the large patch-only capability gradient. The τ-bench policy-versus-reward mismatch instead illustrates a separate, directly observable way that agent behavior changes what a judge is measuring.

### 5.6 Post-submission external-scaffold validation

All eight OpenHands entries met the pre-verdict source threshold (245-250 graded cells each): 1,989 of 2,000 possible agent-issue patches had verifiable graded outcomes. Among 6,708 planned external first judge calls, 17 initially unparseable responses received one retry, ten upstream request errors remained missing, and there were no returned-model identity mismatches or post-retry invalid verdicts. Matching **all three** judges per agent-issue cell leaves 1,981 OpenHands observations; the descriptive Agentless submission contributes 245 separately. The 1,981 matched OpenHands cells include 29 patch prompts truncated at the original 60,000-character cap, but no empty patches or truncated issue descriptions (Appendix F).

**Differential error and the prespecified capability contrasts.** All three judges reject external H1 conditional on issue and execution label (within-cohort Holm p=0.0003 each; 9,999 permutations). This is a global rejection, not a declaration that every pair differs. Raw false-positive rates across the eight configurations span 23.0-81.8% (Gemini), 33.6-78.8% (GPT-5 mini), and 9.4-57.6% (GPT-6 Sol); the primary association uses **task-conditioned** contrasts rather than these raw rates.

| Judge | Resolve rate vs. failed-patch contrast, Spearman | Resolve rate vs. Youden contrast, Spearman | Failed-patch contrast: top minus bottom four |
|---|---:|---:|---:|
| Gemini 3.5 Flash | +0.905 | -0.976 | +27.6 points |
| GPT-5 mini | +0.976 | -0.952 | +21.6 points |
| GPT-6 Sol | +0.952 | -0.881 | +18.4 points |
| Three-judge mean | **+0.944** | **-0.937** | All three positive |

*Table 3: Eight prespecified OpenHands configurations. Agent resolve rates and the three judges' contrasts use the identical 1,981 complete agent-issue cells. The joint eight-agent permutation yields exact two-sided p=4/40,320=0.0000992 for the false-positive contrast and p=16/40,320=0.0003968 for the Youden contrast. Top/bottom differences are descriptive, not independent hypothesis tests.*

Both signs meet the follow-up's **joint directional criterion**. The 200 task/reference-label-conditional verdict shuffles place neither observed average rank correlation inside the simulated null distribution (Monte Carlo p=1/201=0.00498 for each); this check does not manipulate patch content or estimate a causal capability effect. The expected sign does not depend on one particularly favorable submission: dropping each OpenHands configuration in turn leaves mean false-positive correlations of +0.917 to +0.964 and Youden correlations of -0.964 to -0.869 (largest exact p-values 0.00159 and 0.00595). These overlapping subsets are **not** eight independent replications.

| Prespecified OpenHands restriction | Matched cells | Issues | Mean false-positive rho, exact p | Mean Youden rho, exact p |
|---|---|---|---|---|
| Primary, with truncated inputs | 1,981 | 250 | +0.944; 0.000099 | -0.937; 0.000397 |
| Exclude 29 truncated patches | 1,952 | 250 | +0.944; 0.000099 | -0.937; 0.000397 |
| Exclude every initially incomplete issue for all agents | 1,752 | 219 | +0.944; 0.000198 | -0.841; 0.001339 |

*Table 4: The last row removes the union of ten issues with missing public source cells and 22 issues with initially incomplete judge responses (31 distinct issues); an initially unparseable response can be recovered on retry. Network errors remain missing, never negative verdicts.*

The separate Agentless case has a 40.4% execution resolve rate and three-judge complete data on 245 cells; its false-positive rates are 28.8% (Gemini), 49.3% (GPT-5 mini), and 10.3% (GPT-6 Sol). One configuration provides **no Agentless capability slope** or causal estimate of scaffold differences. Figure 6 displays the original mini-SWE-agent and new OpenHands slopes in separate panels because their within-cohort contrast baselines are different.

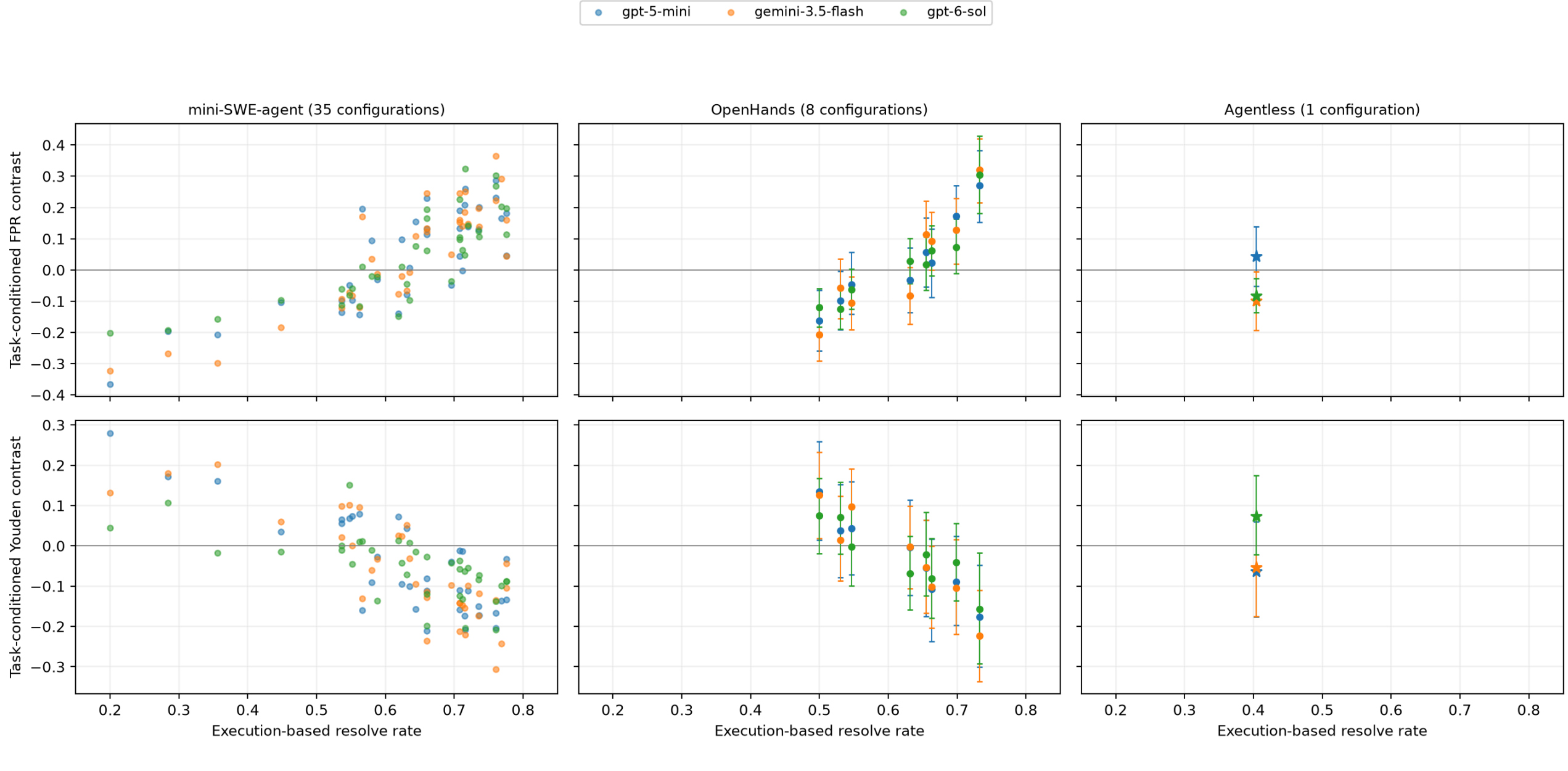

Figure 6. Across the same 250 issues, task-conditioned false-positive contrasts rise and Youden contrasts fall with execution resolve rate in the original mini-SWE-agent cohort (left) and the post-submission OpenHands cohort (middle). The sole Agentless point (right) is descriptive. Each color is one of the same three available judges; the new-cohort intervals resample tasks. Peer sets differ across panels, so vertical levels cannot be read as causal scaffold differences.

**Test-retest anchors and remaining reference uncertainty.** All 300 planned old/new judge pairs on the 100 selected original cells have evaluable verdicts. The old/new same-prompt agreement is 86/100 for GPT-5 mini (95% task-cluster interval 0.789-0.923; Cohen's kappa 0.667), 91/100 for Gemini (0.851-0.961; kappa 0.795) and 87/100 for GPT-6 Sol (0.798-0.938; kappa 0.738). This does not separate temporal provider changes from ordinary decoding variation, nor does it calibrate away external H6. The eight OpenHands configurations span different models, dates, inference budgets and critic settings; one 30B-labelled artifact even contains an inconsistent model-name field. No fixed old/new version lineage exists for this cohort, so H2/H3 release-decision and calibration-transfer findings are **not** replicated here. The same issues are reused, and failing a test does not necessarily imply a semantically incorrect patch. A separately fixed two-human blind audit has 489 candidate patches and 978 blank assignments, but at this revision **no human reviewer has supplied a label**. It cannot support a human-validated claim.

## 6 Practical guidance

The analyses suggest a release procedure that keeps the convenience of a judge while making the decision rest on labels where it matters.

1. **Treat judged differences as screening, not evidence.** A judge-only difference between two versions can be attenuated, biased by differential error, and noisier than a labeled comparison on the same tasks. It is useful to decide which comparisons deserve an audit, not to ship.

2. **Audit the comparison, not the level.** Draw n tasks uniformly at random from the evaluation pool, obtain the reference outcome for *both* versions on each task (execution, tests or expert review), and estimate the paired difference with power-tuned prediction-powered inference (PPI++), using the judge's verdicts on all tasks as the proxy. Because the audit sample is random, the estimator is unbiased for any fixed weight on the judge, however the judge's errors depend on the version, and its interval is asymptotically valid. When the weight is tuned on a small audit, coverage falls below nominal (89.8% at 20 tasks and about 93–94% at 40 on AgentRewardBench and τ-bench), so use Student-t quantiles, treat 40 tasks as a rough planning floor rather than a guarantee, and check coverage for the design at hand or tune the weight on a separate pilot sample.

3. **Size the audit from the comparison's own correlation, and respect the pool.** Let the pool contain N tasks and $S^2 = \Sigma_t (D_t - \bar{D})^2 / (N - 1)$ be the finite-population variance of the task-level differences $D_t = H_{n,t} - H_{o,t}$ (for binary outcomes, $S^2 = N/(N - 1) \cdot (\delta - \bar{D}^2)$), where δ is the fraction of tasks on which the versions' outcomes differ). As an oracle planning approximation — it treats the judge's weight and correlation $\rho_D$ as known — PPI++ has power $1 - \beta$ in a two-sided level-α test of the pool difference against a difference Δ when

$$n \gtrsim \left[\frac{1}{N} + \frac{\Delta^2}{(z_{1-\alpha/2} + z_{1-\beta})^2 S^2 (1 - \rho_D^2)}\right]^{-1},$$

   and the classical audit is the case $\rho_D = 0$. For two versions whose outcomes differ on 25% of tasks and whose success rates differ by Δ = 5 points, testing at α = 0.05 with 80% power in a pool of 300 tasks requires about 217 labeled task pairs without a judge and 202 with a judge whose $\rho_D$ is 0.46, the AgentRewardBench median — a saving of 7%. In a pool of 500 tasks the numbers are 305 and 276. If the target is the difference on future tasks rather than on the pool, the requirement becomes $(z_{1-\alpha/2} + z_{1-\beta})^2 (\delta - \Delta^2)(1 - \rho_D^2)/\Delta^2$, about 613 with the judge and 778 without, which exceeds the size of either pool: a benchmark of a few hundred tasks cannot resolve a 5-point difference between versions this close, with or without a judge. Proposition 3 explains why $\rho_D$ is small for close versions; the judge's main value is to decide which comparisons deserve an audit.

4. **Do not transport calibration from the previous version unless its invariance is checked.** Estimating the judge's sensitivity and specificity on the old version and correcting the new version's judged rate removes attenuation but amplifies any differential error by $1/J$ (Proposition 2), and in our data transport made comparisons worse on every dataset. Transport is exact only when the error rates are the same for both versions, which can be checked only with labels on the new version; if calibration is needed, estimate it on the new version's outputs, which is what the paired audit does.
5. **Keep anchors, but do not treat them as validity checks.** Fixed-anchor monitoring detects changes of the judge. It cannot detect error induced by the new version's outputs, so each release still needs an audit of current outputs.
6. **Report class-conditional judge error by version.** Publishing each version's audited false-positive and false-negative rates next to the judged score makes differential error visible.
7. **Test rather than assume an effect of agent self-report.** An agent's final summary is a potential influence channel, but the randomized intervention in Section 5.4 did not measurably increase average false acceptance. Test input channels and disclosure policies on the specific judging protocol before claiming that hiding a report improves validity.

## 7 Limitations

**Reference standards.** Execution-based labels are not ground truth. SWE-bench tests accept some incorrect patches and reject some correct ones (Yu et al., 2025; Wang et al., 2026; Aleithan et al., 2024), τ-bench rewards check the final database state and can miss policy violations (Zhu et al., 2025), and AgentRewardBench experts agree with each other on 89.3% of doubly annotated test trajectories. If the reference errs more for some agents than for others, part of what we call differential judge error is differential reference error. The reference-aware adjudication in Section 5.2 probes this possibility, but cannot establish a bound without known adjudicator accuracy: it is itself a language model, sees the maintainers' patch, and can err in the opposite direction. The expected-correction analysis additionally assumes shrinkage and interpolation for unsampled vote strata. The post-submission OpenHands follow-up still uses these **execution** labels on the same issues. Its prepared two-reviewer blind packet has no completed human annotations; neither the GPT-5.5 adjudicator nor the three judges are substitutes for independent human correctness labels.

**Judges and access.** The four judges were queried through a gateway to commercial providers with provider-default decoding and one sample per request. Providers can change models, system prompts or decoding without notice; requests were issued over about three days in a random order, so drift adds noise but may still introduce bias if the outage or providers' behavior aligns with some cells. Claude became unavailable before completing SWE-bench and the mechanism experiment, preventing the originally planned four-judge primary analysis; its partial data are descriptive only. The concurrent control arm estimates test-retest agreement across time and prompt conditions, not a pure same-time stochastic error rate. Missing arms were excluded as complete cells per judge; Gemini's five persistently failing requests were upstream errors, and differential missingness cannot be fully ruled out. The three-arm intervention tests adding a short final message to a patch-only input; it does not evaluate full trajectories, report credibility under other instructions, or changes to the judge's model or calibration. In 17 of the 2,100 intended cells, the own and same-agent placebo messages yield identical prompts; those cells cannot distinguish an effect of message content and share a completion cache key. They are retained under the fixed full-sample rule; the net one-vote shifts after replay are disclosed in Section 4.6. The SWE-bench judge sees the repository name, issue and patch, not the full trajectory or the tests; judges that can execute code or read the trajectory may err differently. The judges may have seen SWE-bench issues and reference patches during training, which could raise their accuracy on public tasks without obviously changing its dependence on the agent. On 100 identical mini-SWE-agent patches rejudged for the follow-up, 86-91% of verdicts agree with the originals. The remaining flips cannot be assigned to temporal model drift rather than stochastic decoding or upstream changes, and this selected panel does not bound drift for every external patch style.

**Agents and version pairs.** The SWE-bench agents share one scaffold family, so the version pairs vary models, reasoning settings and scaffold releases within that family; other scaffolds, and private development versions that differ less than public releases, may behave differently. Version pairs are observational: a released upgrade can change several things at once, and pairs share agents, so pair-level results are not independent. τ-bench contributes two agents and AgentRewardBench four, which limits what these replications can say about capability gradients. Eight additional OpenHands-labelled submissions broaden the **observed configuration family** but change underlying models, budgets, critic usage and submission dates together. An artifact labelled 30B has an inconsistent internal model-name field, so even the underlying model sizes should not be interpreted as a controlled treatment. One Agentless submission establishes no within-scaffold slope. All new coding-agent data reuse the original 250 issues, not a fresh task distribution, and do not add pre-identified OpenHands old/new version pairs.

**Statistics.** Some analyses were added after the internal preregistration, in response to an internal, model-assisted review of the frozen plan; they were logged in the private repository before the SWE-bench, τ-bench and mechanism data were read, and we label them throughout. The absence of independent public time-stamping limits the verifiability of this ordering. The AgentRewardBench analyses were not blind: those data are public and were used in the pilot. With 250 tasks, comparisons of close versions have low power, so for version pairs we emphasize interval estimates and minimum detectable effects over counts of significant results. Proposition 3 assumes a single linear proxy and marginally non-differential error; it describes what one binary verdict can offer, not the limit of every judge-based estimator. Replaying the cache after the H5 data lock changed four judge-by-label report-arm acceptance totals by a net one vote each (Section 4.6). The originally collected row-level file was overwritten; the final numbers use the cache-consistent snapshot and the original aggregate effects and byte checksum are retained in the deviation log and checksum manifest, respectively. Canonical cell ordering now makes the seeded analysis independent of asynchronous request-completion order. The external-scaffold protocol was fixed privately **after** the original paper's results were known, albeit before the new cohort's confirmation verdicts. The original arXiv submission had not yet been publicly announced at the follow-up lock; a phrase in the frozen external protocol calling the original results "public" was inaccurate and is corrected here. A private Git commit is not independent public preregistration. Exact eight-agent permutations and task bootstrap intervals condition on the selected submissions; neither treats eight heterogeneous public leaderboard entries as a random sample of future agents.

## 8 Discussion

**Two scales of measurement error.** Holding a judge fixed does not hold its behavior fixed *conditional on what it sees*. With the task and reference label held fixed, the stronger SWE-bench agents' unsuccessful patches receive more positive verdicts from all three available judges; the task-conditioned false-positive contrast grows with agent capability even as the corresponding Youden contrast falls. The same tendency is visible among AgentRewardBench's four agents and, for three judges, τ-bench's two agents, although neither replication establishes a general capability law. A different source of error comes from *which tasks each agent fails*. The share of other agents that solve a task is positively associated with false acceptance on AgentRewardBench, but negatively associated on SWE-bench — even after controlling for agent and source benchmark or repository. H7's predicted positive direction was thus falsified in the primary domain. Neither sign proves that failed trajectories are "near misses": conditioning on observed outcomes selects tasks and potentially erroneous reference labels. The leave-pair-out error split quantifies a prediction residual with reported task/label support, not an identified causal effect of agent identity.

**Transport across a second coding-agent configuration family.** The post-submission OpenHands follow-up holds task identifiers, judges and patch-only prompt fixed while changing the public submission family. Its separate pre-verdict rule is met by all three judges: more successful configurations' failed patches receive higher task-conditioned false acceptance, while their Youden contrasts decline. The result persists when every initially incomplete issue is removed for every OpenHands configuration; the original mini-SWE-agent observation is thus not confined to that named scaffold. But common benchmark issues, mixed models and budgets, an execution-test rather than human correctness reference, and a small nonrandom cohort all limit transport. No OpenHands version-pair release decision, old-to-new calibration ratio, or causal capability effect was identified by this follow-up. The test-retest anchors measure agreement, not a correction for these limitations.

**The release decision is not the leaderboard ranking.** A judge can rank a collection of agents well and still give unjustifiably narrow intervals on the differences that matter for a release. On SWE-bench the judges' Kendall correlations with the reference ranking are 0.71–0.79, yet eight of 33 reference-inconclusive judge-pair comparisons are promoted as significant upgrades. Among the 20 prespecified pairs, nine have a reference-detected improvement and eleven are inconclusive; no significant decline occurs. For τ-bench, by contrast, a judge declares a significant *reverse* ordering on retail despite a nine-point reference gap. AgentRewardBench mostly shows inconclusive judge decisions where the reference detects a difference. A single aggregate accuracy, Brier score or rank correlation cannot stand in for the error on the specific old-to-new comparison, nor for a defensible uncertainty interval.

**The estimand matters as much as the judge.** The τ-bench reversal is compatible with a specific mismatch between the reward's database-state success and the prompt's request for full policy compliance: the agents use tool-and-message turns at very different rates. The stricter judge is not necessarily wrong by a human policy-compliance standard; it is wrong *as a proxy for the published reference reward*. Which standard is appropriate is a product decision that must precede calibration. Agent-specific calibration on new outputs can in principle adjust an aggregate rate for the selected standard, but transporting old-agent error rates without validating invariance cannot do so when habits or patch styles shift.

**A negative randomized mechanism test.** Adding an agent's own final report to the patch did not systematically increase acceptance of failed patches relative to a concurrent patch-only judgment. All three available judges failed the prespecified directional H5 test, even though their patch-only errors differ strongly across agents. Nor did a report from an unrelated issue by the same agent produce an average effect. The randomization identifies an effect of *showing this additional message under this prompt*, not an effect of all self-reports, full trajectories, or agent identity. Agent-specific effects could cancel in the average, and Gemini's exploratory heterogeneity signal should be replicated before using it to motivate a mechanism claim. Together with the opposite H7 gradients across domains, these negative tests caution against treating a plausible narrative about near misses or self-report persuasion as established by observational judge errors.

**Why a large judge budget buys little comparison power.** A prediction-powered paired audit can be valid for an arbitrary judge once its labeling design and weight are controlled, but it need not save many labels. The versions being compared resolve largely the same tasks: median reference outcome correlations are 0.46 on AgentRewardBench and 0.61 for SWE-bench version pairs. The remaining comparison signal lives on discordant tasks. There, the SWE-bench judges' observed label-matched discriminability is lower than on concordant tasks, consistent with — but not a validation of — Proposition 4, since error also depends on the agent. Median gains above one in effective labeled sample size for paired differences are only about half the gains for levels on AgentRewardBench and SWE-bench, and smaller still on τ-bench. On SWE-bench, tuning a proxy narrows an 80-pair audit interval by roughly 5%; on AgentRewardBench it narrows it by 13%. When the target is performance on future tasks, a few hundred benchmark tasks cannot distinguish a five-point version gap with 80% power under the illustrative variances of Section 6. This is a statement about the *design and task pool*, not a universal ceiling on better future evaluators.

**Transport and aggregation are not substitutes for current labels.** Old-version Rogan–Gladen correction amplified the error of 59 of 60 available SWE-bench judge-pair comparisons. Its observed error ratio of 5.20 exceeds the conditional and marginal simulated non-differential references by wide margins, as do the weaker but still adverse AgentRewardBench results. Yet its magnitude is not universal: Claude Opus 5.5 on the partial, matched SWE-bench subset has a much smaller increase. Ensembling the judges cancelled opposing tendencies on τ-bench, but did not eliminate differential error on SWE-bench; tie rules and input probabilities changed the aggregate result. A team should monitor a fixed judge on anchors, screen cheaply with judges, and obtain current paired reference labels before making consequential release decisions.

**Relation to prior work.** Dorner et al. (2025) established limits on the label efficiency of LLM judges; Fiedler (2026) characterized the bias of shared calibration; GAUGE (Bodhwani et al., 2026) found close-pair failures and recalibration problems on $\tau^2$-bench; and Huang (2026) showed that the choice of reference and target alters leaderboard conclusions. Our propositions reuse elementary measurement-error and proxy-inference identities to make the paired-release estimand explicit. The distinct evidence is the preregistered analysis of 20 public agent-version comparisons with execution labels, jointly resampled uncertainty, a randomized additional-input intervention, and an explicitly falsified task-level prediction across domains. Its strength rests on transparent deviations and the quality of its reference labels, not on a claim that model-dependent judge error was previously unknown.

## 9 Conclusion

Using a frozen language-model judge is not enough to make comparisons between versions of an agent valid. On 20 prespecified SWE-bench version pairs, all three available judges showed strong task-conditioned, agent-dependent error; their differential comparison components were detectable in 32 of 60 judge-pair units, and eight judge-only decisions declared an upgrade that execution-based intervals could not establish. In τ-bench, one judge reached a confident decision in the *opposite* direction from the environment reward. Its rejection rationales and the conversations themselves reveal a mismatch between policy compliance and the reference reward, illustrating why a judge's definition of “success” must be aligned with the decision target.

The separately locked post-submission validation finds the same direction of task-conditioned capability/error association in eight OpenHands-labelled configurations, across all three available judges, but on the **same 250 issues** and with configurations that differ in more than capability. It neither establishes a causal explanation nor reproduces the original version-pair release decisions. Independent human verification of the execution reference remains necessary before treating this as a human-correctness result.

Transporting an old agent's error rates is fragile when those rates are not invariant: it increased mean absolute comparison error in all three datasets, by factors from 1.55 to 5.20, with substantial differences among judges. Prediction-powered paired audits, grounded in randomly selected reference-labeled tasks, avoid that bias but often save only a small fraction of labels for close versions. The task-level solvability association even **reverses sign** between AgentRewardBench and SWE-bench; adding the agent's final report to patch-only judging also failed its prespecified directional false-acceptance test for every available judge. No universal story about near misses or self-report contamination explains agent-dependent judge error. A practical evaluation program should state its reference standard, report version- and task-conditioned error, use the judge for screening, and base release decisions on a paired audit of the outputs being compared rather than on frozen-anchor scores or transported calibration alone.

## Broader impact

This study concerns how software teams decide whether a new version of an AI agent is better than the previous one. Its main risk is misuse in the opposite direction from its intent: a reader could take the finding that judges help little for close comparisons as a reason to skip evaluation altogether. Our recommendation is the reverse — audit the comparison with reference labels, and use judges to decide where to audit. The study used only public benchmark data and public agent submissions; no new human subjects or private personal data were collected. Synthetic customer names appear in the public τ-bench trajectories. The model calls were made through a local gateway to commercial model services.

## Data and code availability

The paper's aggregate tables and figures are included here. The code, prompts, prespecified samples, hashes, deviation log and aggregate result tables are archived in a separate **private** research repository, while row-level judge outputs and completion caches are held outside that repository on the author's research machine. Neither is publicly downloadable at posting; this limits independent reproduction without access from the author. The benchmark inputs themselves are public (SWE-bench Verified and the mini-SWE-agent submissions on its leaderboard, τ-bench historical trajectories, and AgentRewardBench), subject to their own licenses and research-use terms; AgentRewardBench-derived material remains subject to its research-use terms. No third-party raw benchmark artifact or completion cache is redistributed with this paper. A rerun also depends on access to the named model services, whose behavior can change over time. The revised paper additionally includes the post-submission OpenHands aggregate tables and comparison figure. The per-submission artifact digests, separately dated pre-verdict protocol, 20 aggregate external files and four aggregate test-retest files are archived in that **private** repository; the graded patch texts, model attempts and SQLite checkpoint remain outside Git. This addition does not make the research archive public or the original private commit an independent timestamp. The 489-case blinded packet has not been scored by either human reviewer; no human audit estimate is reported.

## Use of generative AI

GPT-6 Sol, accessed through Copilot SDK, substantially assisted the literature review, experimental design, code implementation and review, data analysis, the post-submission analysis and figure preparation, and drafting and editing of this manuscript. Claude Opus 5.5 was also used for a separately prompted methodological critique. The internal model-assisted reviews were not external human peer review. These tools are not authors. The primary reference labels were produced by benchmark tests, database-state rewards, or human annotation, not by a writing assistant; every quantitative result reported here is derived from recorded data, an analytic calculation, or an explicitly specified simulation or planning illustration.

## Appendix A: Judge prompts

**SWE-bench system prompt (v1).**

```
You are an expert software engineer reviewing a proposed change to an open-source Python
repository. You will receive a GitHub issue and the patch that an automated coding agent produced
for it.

Decide whether the patch resolves the issue. A patch resolves the issue if, once applied, the
problem described in the issue is fixed in the way the maintainers would expect, the repository's
tests that check this issue would pass, and existing functionality keeps working.

Respond in exactly this format:
<rationale>two to five sentences explaining your judgment</rationale>
<verdict>RESOLVED or NOT_RESOLVED</verdict>
<probability>your probability, between 0 and 1, that the patch resolves the issue</probability>
```

**SWE-bench user message (v1).** The `{self_report_block}` is empty in the primary condition; in the mechanism condition it contains the agent's final message inside `<agent_final_message>` tags. Issues longer than 30,000 characters, patches longer than 60,000 characters and final messages longer than 6,000 characters are truncated with a marker; empty patches are shown as "(the agent submitted an empty patch)".

```
Repository: {repo}

<issue>
{problem_statement}
</issue>

<patch>
{patch}
</patch>
{self_report_block}
```

**τ-bench system prompt (v1).** The `{policy}` placeholder receives the domain policy distributed with τ-bench. The user message is the rendered conversation: user and agent turns, each tool call with its arguments, and each tool result truncated to 4,000 characters; the agent's own system prompt is omitted.

```
You are evaluating an AI customer-service agent. The agent talks with a user, uses tools to read
and modify a database, and must follow the domain policy below.

Decide whether the agent successfully completed the user's request. A conversation is successful
only if the agent made every database change the user was entitled to and asked for, made no
incorrect or unauthorized changes, followed the policy, and correctly provided any information the
user asked for.

<policy>
{policy}
</policy>

Respond in exactly this format:
<rationale>two to five sentences explaining your judgment</rationale>
<verdict>SUCCESS or FAILURE</verdict>
<probability>your probability, between 0 and 1, that the agent successfully completed the
request</probability>
```

## Appendix B: SWE-bench version pairs

Reference resolve rates are computed on the 250 confirmation tasks from the recorded test outcomes. Discordance is the fraction of tasks on which the two versions' outcomes differ; it bounds the variance of the paired difference. Main judge comparisons use the common agent-task cells on which the selected judges responded, so some paired differences in Section 5.2 differ slightly from these full-reference values because seven upstream request failures remain.

| # | Type | Old version | New version | Old rate | New rate | Difference | Discordance |
|---|---|---|---|---:|---:|---:|---:|
| 1 | model | o3 (2025-04-16), mini-SWE-agent v1.0.0 | GPT 5 (2025-08-07) (medium), mini-SWE-agent v1.7.0 | 0.580 | 0.660 | +0.080 | 0.176 |
| 2 | model | GPT 5 (2025-08-07) (medium), mini-SWE-agent v1.7.0 | GPT 5.1 (2025-11-13) (medium), mini-SWE-agent v1.15.0 | 0.660 | 0.644 | -0.016 | 0.128 |
| 3 | model | GPT 5.1 (2025-11-13) (medium), mini-SWE-agent v1.15.0 | GPT 5.2 (2025-12-11), mini-SWE-agent v1.17.2 | 0.644 | 0.708 | +0.064 | 0.176 |
| 4 | configuration | GPT 5.2 (2025-12-11), mini-SWE-agent v1.17.2 | GPT 5.2 (2025-12-11) (high), mini-SWE-agent v1.17.2 | 0.708 | 0.716 | +0.008 | 0.096 |
| 5 | variant | GPT 5.1 (2025-11-13) (medium), mini-SWE-agent v1.15.0 | GPT 5.1 Codex (medium), mini-SWE-agent v1.16.0 | 0.644 | 0.660 | +0.016 | 0.136 |
| 6 | scaffold | GPT 5.2 (2025-12-11) (high), mini-SWE-agent v1.17.2 | GPT 5.2 (high), mini-SWE-agent v2.0.0 | 0.716 | 0.760 | +0.044 | 0.132 |
| 7 | model | o4-mini (2025-04-16), mini-SWE-agent v1.0.0 | GPT 5 mini (2025-08-07) (medium), mini-SWE-agent v1.7.0 | 0.448 | 0.624 | +0.176 | 0.240 |
| 8 | scaffold | GPT 5 mini (2025-08-07) (medium), mini-SWE-agent v1.7.0 | GPT 5 mini, mini-SWE-agent v2.0.0 | 0.624 | 0.564 | -0.060 | 0.204 |
| 9 | model | Claude 4 Sonnet (20250514), mini-SWE-agent v1.0.0 | Claude 4.5 Sonnet (20250929), mini-SWE-agent v1.13.3 | 0.636 | 0.712 | +0.076 | 0.172 |
| 10 | scaffold+configuration | Claude 4.5 Sonnet (20250929), mini-SWE-agent v1.13.3 | Claude 4.5 Sonnet (high), mini-SWE-agent v2.0.0 | 0.712 | 0.708 | -0.004 | 0.148 |
| 11 | model | Claude 4 Opus (20250514), mini-SWE-agent v1.0.0 | Claude 4.5 Opus (20251101) (medium), mini-SWE-agent v1.16.0 | 0.696 | 0.732 | +0.036 | 0.164 |
| 12 | scaffold+configuration | Claude 4.5 Opus (20251101) (medium), mini-SWE-agent v1.16.0 | Claude 4.5 Opus (high), mini-SWE-agent v2.0.0 | 0.732 | 0.768 | +0.036 | 0.140 |
| 13 | model | Claude 4.5 Opus (high), mini-SWE-agent v2.0.0 | Claude 4.6 Opus, mini-SWE-agent v2.0.0 | 0.768 | 0.776 | +0.008 | 0.120 |
| 14 | model | Gemini 2.5 Pro (2025-05-06), mini-SWE-agent v1.0.0 | Gemini 3 Pro Preview (2025-11-18), mini-SWE-agent v1.15.0 | 0.548 | 0.720 | +0.172 | 0.244 |
| 15 | model | GLM 4.5 (2025-08-22), mini-SWE-agent v1.9.1 | GLM 4.6 (T=1), mini-SWE-agent v1.17.1 | 0.560 | 0.552 | -0.008 | 0.208 |
| 16 | model | GLM 4.6 (T=1), mini-SWE-agent v1.17.1 | GLM 5 (high), mini-SWE-agent v2.0.0 | 0.552 | 0.736 | +0.184 | 0.280 |
| 17 | model | Kimi K2 Thinking, mini-SWE-agent v1.17.2 | Kimi K2.5 (high), mini-SWE-agent v2.0.0 | 0.628 | 0.708 | +0.080 | 0.168 |

| # | Type | Old version | New version | Old rate | New rate | Difference | Discordance |
|---|---|---|---|---|---|---|---|
| 18 | model | MiniMax M2, mini-SWE-agent v1.17.0 | MiniMax M2.5 (high), mini-SWE-agent v2.0.0 | 0.616 | 0.776 | +0.160 | 0.224 |
| 19 | scaffold+configuration | DeepSeek V3.2 Reasoner, mini-SWE-agent v1.17.1 | DeepSeek V3.2 (high), mini-SWE-agent v2.0.0 | 0.588 | 0.712 | +0.124 | 0.228 |
| 20 | variant | Devstral Small (2512), mini-SWE-agent v1.17.2 | Devstral (2512), mini-SWE-agent v1.17.2 | 0.536 | 0.536 | +0.000 | 0.248 |

## Appendix C: Simulation study of the diagnostics

To read the empirical diagnostics against known truths, we simulated six agents of evenly spaced ability (−1 to 1 on the logit scale) attempting the same 250 tasks, with task difficulty drawn from a standard normal distribution. The judge accepts a failed trajectory with probability logistic($-1.7 + c_t + \gamma_F a$) and a successful one with probability logistic($1.7 - c_t + \gamma_T a$), where $a$ is the agent's centered ability and $c_t$ a task-level judging difficulty that shares half of its variance with task difficulty (standard deviation 1 except in the null scenario, where it is 0), so that agents which fail different tasks face different marginal error rates even when the judge has no agent-specific error. Each scenario was simulated 200 times (`scripts/simulations.py`; 499 permutations and 300 bootstrap draws per run); pair diagnostics refer to the five pairs of adjacent agents.

| Scenario | $\gamma_F$ | $\gamma_T$ | H1 rejects (prereg.) | H1 rejects (label-specific) | Decision disagreement | Transport ratio | Mean agent-specific error |
|---|---|---|---|---|---|---|---|
| Null, no task-level judging difficulty | 0 | 0 | 5.5% | 7.5% | 35.2% | 1.10 | 0.000 |
| Task mix only (conditionally non-differential) | 0 | 0 | 5.0% | 5.0% | 36.8% | 1.17 | 0.000 |
| FPR rises with ability | 0.5 | 0 | 82.0% | 59.5% | 34.7% | 1.57 | +0.014 |
| FPR falls with ability | −0.5 | 0 | 0.0% | 59.5% | 40.6% | 1.01 | −0.014 |
| TPR rises with ability | 0 | 0.5 | 50.0% | 33.0% | 31.8% | 1.58 | +0.009 |

Rejection rates are Monte Carlo estimates from 200 runs; near 5% their standard error is about 1.5 percentage points, so the 95% Monte Carlo interval for the null rates spans roughly ±3 points. Four conclusions follow. First, neither H1 test shows a detectable size distortion when marginal error rates differ only because of task mix (5.0% for both; 5.5% and 7.5% in the null scenario), so a rejection points to agent-specific error. Second, the preregistered statistic has no power when false acceptance falls with ability, whereas the label-specific statistic detects both directions, at some cost in the directions the preregistered statistic favors; reporting both is therefore informative. Third, a third or more of judge-only release decisions disagree with reference-label decisions even for a perfectly non-differential judge, because attenuation turns significant differences into inconclusive ones; disagreement rates must be compared with a null reference rather than with zero. Fourth, the transport-to-naive error ratio exceeds 1 under both null scenarios, so the preregistered H3 criterion (lower bound above 0.5) does not by itself indicate differential error; the ratio rises clearly above its null value when error rates change with ability in the direction that aligns with the true differences, and not when they change in the opposite direction, because then differential error and attenuation add up in the naive comparison, which becomes as wrong as the transported one. The mean agent-specific error is centered on zero in both null scenarios and has the sign of the simulated effect otherwise; with six agents, other agents share each task and label almost always, so these runs do not test the sparse settings of AgentRewardBench (four agents) or τ-bench (two agents, where the decomposition is undefined). With effects of this size, single adjacent pairs rarely show significant agent-specific error (5.4%–7.0% of pairs across scenarios), which is why we pool pairs and report interval estimates.

## Appendix D: Adjudicator prompt

The adjudicator (Section 5.2) receives the maintainers' reference patch and test changes, which the judges never see, and neither the test outcome nor any judge verdict. Texts longer than 20,000 characters (40,000 for the candidate patch) are truncated with a marker.

**System prompt (v1).**

```
You are a senior software engineer auditing how patches submitted for GitHub issues were graded.
You will receive an issue from an open-source Python repository, the reference patch that the
maintainers merged, the test changes the maintainers made to verify the fix together with the names
of the tests that must pass, and a candidate patch written by someone else.

Decide whether the candidate patch correctly resolves the issue. A candidate can be correct even if
it differs from the reference patch, as long as it fixes the behavior described in the issue and
satisfies the intent of the maintainers' tests. A candidate is incorrect if it does not fix the
issue, fixes it only partly, breaks behavior that the issue or the tests imply, or is empty.

Some tests check details that the issue does not specify, such as an exact error message, a new
function or parameter name, or a particular output format. If the candidate fixes the issue as
described but would fail the tests only because of such details, answer CORRECT and report that the
tests are overspecified.

Respond in exactly this format:
<rationale>two to five sentences explaining your judgment</rationale>
<verdict>CORRECT or INCORRECT</verdict>
<tests_overspecified>YES or NO</tests_overspecified>
<probability>your probability, between 0 and 1, that the candidate patch correctly resolves the
issue</probability>
```

**User message (v1).** `{fail_to_pass}` lists the tests that the reference patch makes pass, one per line.

```
Repository: {repo}

<issue>
{problem_statement}
</issue>

<reference_patch>
{gold_patch}
</reference_patch>

<maintainer_test_changes>
{test_patch}
</maintainer_test_changes>

<tests_that_must_pass>
{fail_to_pass}
</tests_that_must_pass>

<candidate_patch>
{patch}
</candidate_patch>
```

## Appendix E: Examples of judge rationales (τ-bench)

Verbatim rationales of GPT-6 Sol for conversations that the environment reward scores as successful (reference label 1) and that the judge rejected. They were drawn at random from the retail domain (seed 20260924) as part of the exploratory analysis in Section 5.3.

- *Claude 3.5 Sonnet, retail task 78, trial 3.* "The agent authenticated Yara, checked the relevant order statuses, obtained confirmation, and completed all three requested database changes correctly. However, it repeatedly sent user-facing text and a tool call in the same turn, contrary to the policy requiring those to be separate. Because success requires following the policy as well as completing the changes, the conversation is not successful."
- *Claude 3.5 Sonnet, retail task 67, trial 3.* "The agent authenticated the user and reported the $829.43 payment shown for order #W6729841. However, it sent customer-facing messages and made tool calls in the same turns, which the policy explicitly prohibits. Because the conversation did not fully follow policy, it is not successful."
- *GPT-4o, retail task 95, trial 2.* "The agent authenticated the user, identified the correct delivered order, obtained confirmation, and submitted the requested laptop exchange. However, the policy specifically requires reminding the customer to confirm they have provided all items to be exchanged. Asking whether the listed details were correct did not explicitly cover that requirement."

In all three cases the judge acknowledges that the user's request was carried out — which is what the reward checks — and rejects the conversation for a procedural requirement of the policy.

## Appendix F: Provenance and coverage of the post-submission scaffold follow-up

The original 250-issue confirmation sample has SHA-256 `a19f1a1efb941b84205bb3690ab4a735796420ec494da19f367ea9ceb4e0a1bc`. The original SWE-bench patch-only system and user prompts have SHA-256

f1008fa2e382dca060b69d5aa100a7022830962ce7f1533cb1bd6f971f9e7eca and f0a91eb02fdee37788bafedf89a94916797eb6ed93a45e610bbdc3d6e5821533. Public submission IDs, prediction digests, execution reports and graded-diff source manifests are pinned to swe-bench/experiments commit 40f164d5. The fixed post-original-results protocol and source design were committed privately at ce8c6a1 on 2026-09-28; the design JSON digest is 0b039a7b29bc3d5e23bd9c1fcf4bed8b52bbc5c664ec4d1113b8537ed039d336. None of these private hashes constitutes public preregistration, and the original paper had not yet been publicly announced at the follow-up lock.

| Public submission ID | Graded / 250 | Three-judge complete | Successful / complete |
|---|---|---|---|
| 20241029_OpenHands-CodeAct-2.1-sonnet-20241022 | 247 | 247 | 135 / 247 |
| 20250203_openhands_4x_scaled | 249 | 247 | 156 / 247 |
| 20250415_openhands | 250 | 249 | 165 / 249 |
| 20250520_openhands_devstral_small | 245 | 244 | 122 / 244 |
| 20250716_openhands_kimi_k2 | 250 | 249 | 163 / 249 |
| 20250805_openhands-Qwen3-Coder-30B-A3B-Instruct | 250 | 249 | 132 / 249 |
| 20250805_openhands-Qwen3-Coder-480B-A35B-Instruct | 250 | 249 | 174 / 249 |
| 20250807_openhands_gpt5 | 248 | 247 | 181 / 247 |
| **OpenHands total** | **1,989** | **1,981** | **1,228 / 1,981** |
| 20241028_agentless-1.5_gpt4o (descriptive) | 247 | 245 | 99 / 245 |

*Table F1: Graded patches were verified against their public evaluation-log diffs where available. No absent execution report was coded as a failure; no mismatched patch was silently substituted. The eight OpenHands configurations have 240 issues in common before new judge errors (238 when Agentless is included). "Successful" means the benchmark's test-based reference outcome, not a human adjudication. Submission labels need not perfectly describe the underlying model; one sampled 30B-labelled artifact names a 480B model.*

The new external attempt CSV (6,725 rows including 17 permitted parse retries) has SHA-256 03e8e6edab71d5efd6138641d72247aec9f060396bf9fd1e05b209d861e20b1e. The 100 mini-anchor cells were each rejudged by three models (300 first attempts and two parse retries); their raw attempt CSV has SHA-256 a9837913936094c379f0f7482e4b49ec11de146ba912b73e04fdaab130928647. The combined off-Git SQLite attempt checkpoint has 7,027 records and SHA-256 7abf87df8e09c3bcd8c26c8ed2c54ee8ac4d435cbfe68b5bd612088658bd8f39. All 20 external and four anchor aggregate JSON/CSV outputs reproduced byte-for-byte from these snapshots in a second analysis, and both exact H6 p-values were independently enumerated from committed aggregate rates and contrasts. The private preregistration/HASHES.txt inventories each aggregate file's digest. Because raw benchmark patches and model outputs are not redistributed, these checks do not constitute an independently reproducible public dataset.